\documentclass[10pt,twocolumn,letterpaper]{article}

\usepackage{abstract}
\usepackage{cvpr}              
\usepackage[dvipsnames]{xcolor}

\usepackage[utf8]{inputenc} 
\usepackage[T1]{fontenc}    
\usepackage{url}            
\usepackage{booktabs}       
\usepackage{amsfonts}       
\usepackage{nicefrac}       
\usepackage{microtype}      
\usepackage{xcolor}         
\usepackage{tabularx,colortbl}
\usepackage{graphicx}
\usepackage{subcaption}
\usepackage{multirow}
\usepackage[title]{appendix}

\usepackage{pifont}
\newcommand{\xmark}{\ding{55}}
\newcommand{\NA}{---}
\newcommand{\parahead}[1]{\noindent\textbf{#1}:\ }

\definecolor{cvprblue}{rgb}{0.21,0.49,0.74}
\usepackage[pagebackref,breaklinks,colorlinks,citecolor=cvprblue]{hyperref}

\def\paperID{11676} 
\def\confName{CVPR}
\def\confYear{2024}

\newcommand{\shortname}{DiVa-360\xspace}
\newcommand{\brics}{BRICS\xspace}
\newcommand{\bricsfull}{Brown Interaction Capture System\xspace}
\newcommand{\bricsfullu}{\underline{B}\underline{r}own \underline{I}nteraction \underline{C}apture \underline{S}ystem\xspace}
\newcommand{\papertitle}{The Dynamic Visual Dataset for Immersive Neural Fields}
\title{\shortname: \papertitle}

\author{Cheng-You Lu$^1$\thanks{Equal Contribution} \and Peisen Zhou$^1$\footnotemark[1] \and Angela Xing$^1$\footnotemark[1] \and Chandradeep Pokhariya$^2$ \and Arnab Dey$^3$\thanks{Work was done while at Brown University} \and Ishaan Nikhil Shah$^2$ \and Rugved Mavidipalli$^1$ \and Dylan Hu$^1$ \and Andrew I. Comport$^3$ \and Kefan Chen$^1$ \and Srinath Sridhar$^1$\and \\
$^1$Brown University \quad $^2$IIIT Hyderabad  \quad $^3$I3S-CNRS/Université Côte d’Azur\\
\href{https://ivl.cs.brown.edu/research/diva}{ivl.cs.brown.edu/research/diva}
}
\begin{document}

\twocolumn[
\maketitle
\begin{center}
  \centering
  \captionsetup{type=figure}
  \vspace{-0.35in}
  \includegraphics[width=1.0\linewidth]{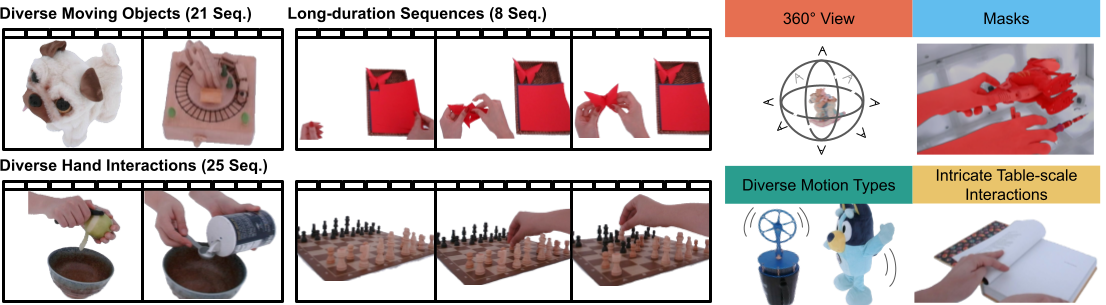}
  \vspace{-0.24in}
  \captionof{figure}{
\shortname is a \textbf{real-world} 360$^\circ$ multi-view visual dataset of dynamic tabletop scenes captured using a customized low-cost capture system consisting of 53 cameras.
\shortname contains 21 diverse moving object sequences, 25 hand-object interaction sequences, and 8 long-duration sequences (2-3 mins).
\shortname provides (1)~360$^\circ$ coverage of dynamic scenes, (2)~foreground-background segmentation masks,
and (3)~diverse table-scale scenes with intricate motions. 
\shortname aims to facilitate research in dynamic long-duration neural fields.
  }
  \label{fig:teaser}
  \vspace{-0.05in}
\end{center}
]
\saythanks

\begin{abstract}
Advances in neural fields are enabling high-fidelity capture of the shape and appearance of dynamic 3D scenes.
However, their capabilities
lag behind those offered by conventional representations such as 2D videos because of algorithmic challenges and the lack of large-scale multi-view real-world datasets.
We address the dataset limitation with \shortname, a real-world 360$^\circ$ \underline{d}ynam\underline{i}c \underline{v}isu\underline{a}l dataset that contains synchronized high-resolution and long-duration multi-view video sequences of table-scale scenes captured using a customized low-cost system with 53 cameras.
It contains 21 object-centric sequences categorized by different motion types, 25 intricate hand-object interaction sequences,
and 8 long-duration sequences for a total of 17.4~M image frames.
In addition, we provide foreground-background segmentation masks, synchronized audio, and text descriptions.
We benchmark the state-of-the-art dynamic neural field methods on \shortname and provide insights about existing methods and future challenges on long-duration neural field capture.


%
\end{abstract}

\begin{figure*}[htb!] 
\centering
\vspace{-5mm}
  \includegraphics[width=\textwidth]{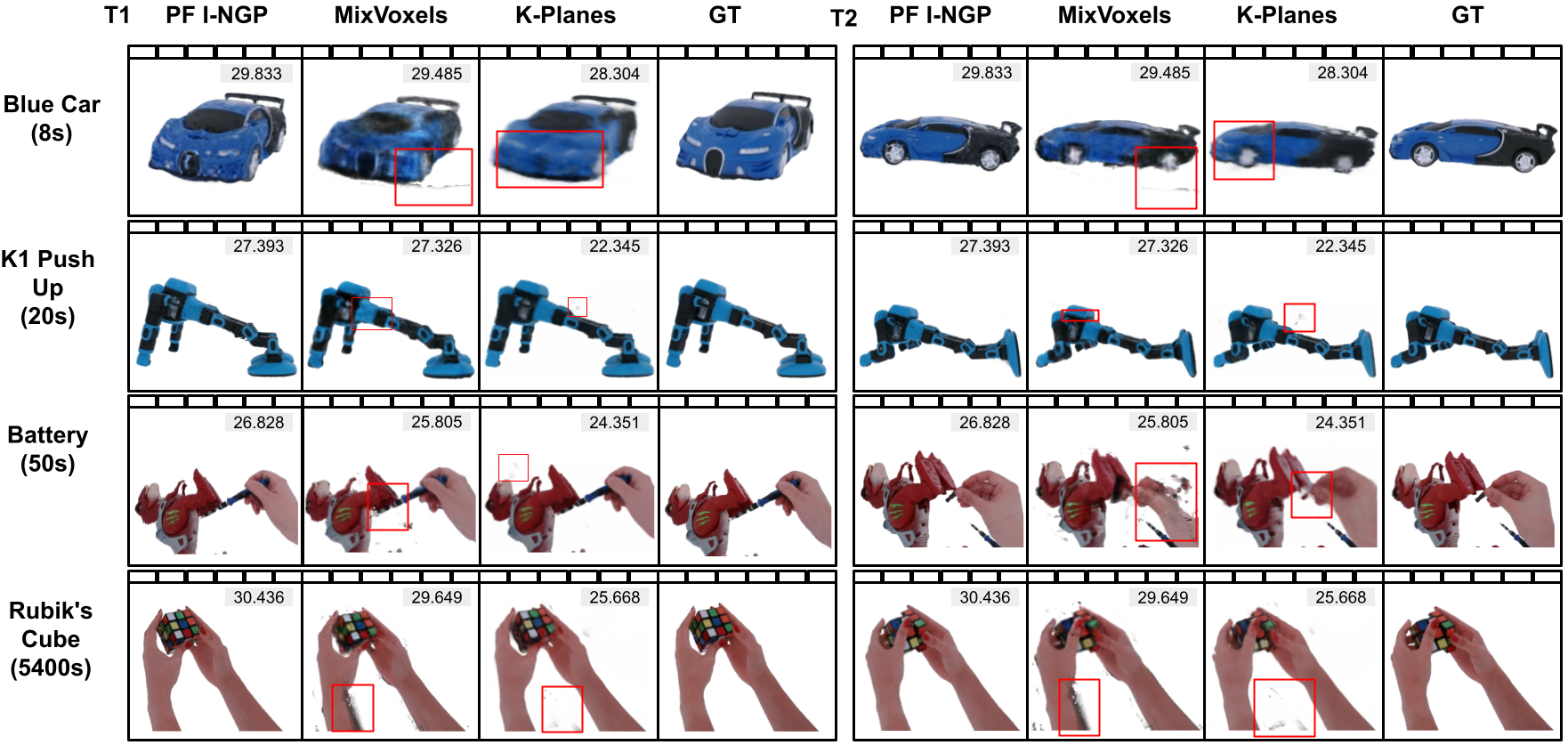}
  \vspace{-3mm}
  \caption{\shortname provides multi-view dynamic sequences for dynamic neural field methods.
  The dataset contains a variety of object and motion types.
  Here, we showcase reconstruction results across time steps from PF I-NGP~\cite{mueller2022instant}, MixVoxels~\cite{wang2023mixed}, and K-Planes~\cite{kplanes_2023} trained on our dataset. Surprisingly, the rendering results of PF I-NGP, a method that does not directly utilize temporal information from adjacent frames, are better than those of MixVoxels and K-Planes. MixVoxels struggles with complex motion data, such as hands, while K-Planes suffers from floaters in the background.
  We demonstrate more visualization results in the supplementary Section 2.
  }
  \label{fig:teaser2}
\end{figure*}

\section{Introduction}
\label{sec:intro}
Neural fields~\cite{xie2022_neuralfields}, or neural implicit representations, 
have recently emerged as useful representations in computer vision, graphics, and robotics~\cite{xie2022_neuralfields,tewari2022advances} for
capturing properties such as radiance~\cite{mildenhall2020nerf,barron2022mipnerf360,barron2021mipnerf,mueller2022instant,kerbl3Dgaussians}, shape~\cite{Park_2019_CVPR,yariv2021volume,neus2,oechsle2021unisurf,mescheder2019occupancynetworks,li2021nsff}, and dynamic motion~\cite{wang2023mixed,dynerf,xian2021space,Gao-ICCV-DynNeRF,Lombardi:2019,broxton2020immersive,park2021hypernerf,luiten2023dynamic,wu20234dgaussians}. 
Their high fidelity, continuous representation, and implicit compression~\cite{dupont2021coin} properties make them attractive as immersive digital representations of our dynamic world.


However, despite their popularity, neural fields remain less capable than conventional representations for representing dynamic scenes.
Consider this: we can easily watch hours-long 2D videos, a task
that cannot yet be achieved efficiently with 3D neural fields due to long training times~\cite{dynerf, kplanes_2023, attal2023hyperreel, wang2023mixed, li2022streaming, nerfplayer, hexplane, kerbl3Dgaussians, wu20234dgaussians, luiten2023dynamic}.
We believe that \textbf{large-scale, real-world} datasets of dynamic scenes with associated benchmarks are essential for continued progress in this problem.
%
While some 
real-world dynamic datasets  exist~\cite{dynerf, yoon2020novel, broxton2020immersive, li2021nsff, tancik2022blocknerf, park2021hypernerf, objaverse, yan2023nerf, VRNeRF, lu2023large}, they are limited to room-scale scenes or specific categories like humans~\cite{isik2023humanrf, tao2021function4d, liu2023hosnerf, peng2021neural, ionescu2013human3, Joo_2015_ICCV}, or are captured with monocular or forward-facing cameras that do not always provide sufficient multi-view cues for immersive reconstruction~\cite{gao2022dynamic,li2021nsff,park2021hypernerf, lu2023large}.
Furthermore, most of the sequences in these datasets~\cite{dynerf, yoon2020novel, li2021nsff, park2021hypernerf, objaverse, yan2023nerf, lu2023large} are short, often less than 15 seconds, limiting their use for building methods that capture long-duration scenes.

To address these limitations, we present \textbf{\shortname}, a real-world \underline{d}ynam\underline{i}c \underline{v}isu\underline{a}l dataset that contains synchronized high-resolution long-duration table-scale sequences captured by a 360$^\circ$ multi-view camera system (see \Cref{fig:teaser}).
Our dataset includes high-resolution (1280$\times$720), high-framerate (120~FPS), and up to 3 mins long videos captured simultaneously from 53 RGB cameras spanning 360$^\circ$ volume within the capture space.
We provide 46 dynamic sequences, including 21 object-centric sequences categorized by different motion types and 25 hand-object interaction sequences with human routine activities
and 8 long duration dynamic sequences. 
In total, \shortname dynamic dataset contains \textbf{17.4~M} image frames of 53 dynamic scenes over \textbf{2738} seconds. 
We also provide foreground-background segmentation masks, accompanying audio data from microphones, and detailed text descriptions of the activity observed.



Capturing such large-scale data requires advances in capture systems, as well as benchmarking metrics.
We have built a new low-cost capture system called \textbf{\brics (\bricsfullu)} which is designed to capture synchronized, high-framerate, and high-fidelity data.
%
In addition, we propose standardized metrics for reconstruction quality and runtime, and compare baseline methods on these metrics~\cite{mueller2022instant, wang2023mixed, kplanes_2023}.
We perform a systematic analysis of current methods and characterize their performance on different sequence durations, image resolutions, motion types, and viewpoints.
Surprisingly, we observe that methods that model each frame in a dynamic sequence without directly using temporal information~\cite{mueller2022instant} outperform state-of-the-art dynamic methods~\cite{wang2023mixed, kplanes_2023} in terms of reconstruction quality and even training speed (see \Cref{fig:teaser2}). 
In addition, existing methods~\cite{wang2023mixed, kplanes_2023} are biased toward moving objects' shapes, thereby losing high-frequency information and fine-grained details.
Finally, existing state-of-the-art methods~\cite{kplanes_2023,wang2023mixed} prefer different amounts of temporal information, with one~\cite{kplanes_2023} starting to outperform another~\cite{wang2023mixed} after acquiring more temporal information.
%
To summarize, we make the following contributions:
\begin{itemize}
    \item \textbf{\brics}: A low-cost capture system specifically designed for 360$^{\circ}$ capture of table-scale dynamic scenes with 53 synchronized RGB cameras. 
    \item \textbf{\shortname Dataset}: The largest dataset (17.4~M frames) for dynamic neural fields with 21 object-centric sequences categorized by different motion types, 25 hand-object interaction sequences including routine human activities, and 8 long-duration dynamic sequences.
    \item \textbf{Benchmark \& Analysis}: We benchmark the dataset with state-of-the-art methods and enable a better understanding of the current state of dynamic neural fields.
\end{itemize}
We believe our work can help the community take a leap from the current focus on short dynamic videos toward a more holistic understanding of longer dynamic scenes. 

\begin{table}[t!]
    \centering
    \scalebox{0.7}{
    \begin{tabular}{llccccccc}
    \toprule
    \textbf{Dataset} & \textbf{Real} & \textbf{Mask} & \textbf{360$^\circ$ view} & \textbf{Multiview} & \textbf{Object}  & \textbf{Scene} \\
    \midrule
    {HyperNeRF~\cite{park2021hypernerf}} & \checkmark   & \xmark & \xmark & \xmark & \xmark & \checkmark \\
    {OMMO~\cite{lu2023large}} & \checkmark  & \xmark & \xmark & \xmark & \xmark & \checkmark \\
     {Block-NeRF~\cite{tancik2022blocknerf}} & \checkmark  & \xmark & \xmark & \checkmark & \xmark & \checkmark \\
     {Eyeful Tower~\cite{VRNeRF}} & \checkmark  & \xmark & \xmark & \checkmark & \xmark & \checkmark \\
    {DyNeRF~\cite{dynerf}}  & \checkmark  & \xmark & \xmark & \checkmark & \xmark & \checkmark  \\
    {ILFV~\cite{broxton2020immersive}} & \checkmark & \xmark & \xmark & \checkmark & \xmark & \checkmark \\
    {Deep3DMV~\cite{lin2021deep}} & \checkmark & \xmark & \xmark  & \checkmark & \xmark & \checkmark \\
    {NeRF-DS~\cite{yan2023nerf}} & \checkmark  & \checkmark & \xmark & \checkmark & \xmark & \checkmark \\
    {NDSD~\cite{yoon2020novel}} & \checkmark  & \checkmark & \xmark & \checkmark & \xmark & \checkmark \\
    \midrule
    {D-NeRF~\cite{pumarola2020d}} & \xmark & \NA & \checkmark & \xmark & \checkmark & \xmark \\
    \midrule
    {Objaverse~\cite{objaverse}} & \xmark & \NA & \checkmark  & \checkmark & \checkmark & \checkmark \\
    \rowcolor{yellow}
    {\shortname} & \checkmark & \checkmark & \checkmark  & \checkmark & \checkmark & \checkmark   \\ 
    \bottomrule
  \end{tabular}
  }
  \caption{We compare featured properties of our \shortname~with other object-centric and scene-centric datasets. \shortname is a unique dataset that contains real-world 360$^\circ$ multiview object-centric and scene-centric data with foreground-background masks.}
  \label{tab:dataset_comparison}
 \end{table}

\section{Related Work}
\label{sec:relatedwork}
%
%
\parahead{Neural Fields}
Neural fields, or coordinate-based implicit neural networks, have generated considerable interest in computer vision~\cite{xie2022_neuralfields} because of their ability to represent geometry~\cite{Occupancy, Park_2019_CVPR, chen2018implicit_decoder} and appearance~\cite{mildenhall2020nerf, Lombardi:2019, NEURIPS2019_b5dc4e5d}.
Neural radiance fields (NeRF)~\cite{mildenhall2020nerf} and its variants~\cite{barron2021mipnerf,yariv2021volume,oechsle2021unisurf,li2023neuralangelo,wang2023mixed} uses a multilayer perceptron (MLP) to model density and color, leading to photorealistic novel view synthesis and 3D reconstruction.
Since the training cost of NeRFs is high, several methods have tried to address this limitation~\cite{mueller2022instant,yu_and_fridovichkeil2021plenoxels,Tensorf,kerbl3Dgaussians}.
Naturally, some approaches have also turned their focus towards dynamic neural fields~\cite{wang2023mixed,dynerf,park2021hypernerf,li2021nsff,park2021nerfies,pumarola2020d,Gao-ICCV-DynNeRF,Lombardi:2019,kplanes_2023,hexplane,nerfplayer, wu20234dgaussians,xu20234k4d,luiten2023dynamic}.
However, these methods have thus far been limited to only brief sequences, partly as a result of the unavailability of long-duration datasets.
Our work enables further research in long-duration dynamic neural field research with a more comprehensive and richer dataset with long sequences.

%

\parahead{Multi-Camera Capture Systems}
\label{sec:relwork_stages}
Capturing multi-view data with high resolution and framerate requires specialized hardware and software systems.
The earliest multi-camera capture systems were extensions of stereo cameras to 5--6 cameras~\cite{kanade1995development}, which were later extended to capture a hemispherical volume~\cite{kanade2007virtualized} with up to 50 cameras for 3D and 4D reconstruction using non-machine learning techniques~\cite{theobalt2004combining}.
The focus of most existing multi-camera capture systems has been on room-scale scenes for human or environment capture~\cite{joo2015panoptic,yu2020humbi}.
While some table-scale datasets exist, notably for hand interaction capture~\cite{zimmermann2019freihand,brahmbhatt2019contactdb}, they have only a limited number of cameras.
In contrast, our \brics system is specially designed for dense 53-view visual capture of table-scale scenes, and our sequences showcase intricate interactions (\eg,~small tool use) in high fidelity.

\parahead{Datasets for Dynamic Neural Fields}
%
While plenty of datasets exist for NeRF methods~\cite{dtu,Knapitsch2017,mildenhall2020nerf,barron2022mipnerf360,mildenhall2019llff,yao2020blendedmvs,reizenstein21co3d,wu2023omniobject3d,deluigi2023scannerf,jeong2022perfception} their focus has been on static scenes.
%
For dynamic scenes, numerous datasets such as DyNeRF~\cite{dynerf}, NDSD~\cite{yoon2020novel}, ILFV~\cite{broxton2020immersive}, NeRF-DS~\cite{yan2023nerf}, and Deep3DMV~\cite{lin2021deep}  exist, but they are limited to only a short duration (\textasciitilde 15s), or have only forward-facing cameras preventing them from enabling 360$^\circ$ capture.
BlockNeRF~\cite{tancik2022blocknerf} provides street view videos incorporating dynamic elements but lacks a focus on objects, and does not provide many views.
This creates fleeting scenes that do not encompass full 360$^\circ$ camera coverage.

Eyeful Tower~\cite{VRNeRF} provides dynamic data up to 2000~s long, but the framerate is less than 4~FPS.
Monocular videos of human faces~\cite{park2021hypernerf, park2021nerfies}, human activities~\cite{li2021nsff}, or outdoor scene~\cite{lu2023large} have been used for neural field reconstructions, but a single camera restricts visibility resulting in low effective multi-view factors (EMF)\cite{gao2022dynamic}.
While datasets like Objaverse~\cite{objaverse} and SAPIEN~\cite{xiang2020sapien} provide articulated objects, they are not sourced from the real world.
Our dataset stands out by offering a 360$^\circ$ view of real-world (non-synthetic) long dynamic sequences with objects and hand-object interaction captured by 53 synchronized cameras (see Table~\ref{tab:dataset_comparison} and Table~\ref{tab:dataset_stats}). Furthermore, each sequence is accompanied by foreground-background segmentation masks. 
Hence, we do not need to worry about the domain gap, the influence from the background, and the insufficient multiview cues.


 \begin{table}[t!]
   \centering
  \scalebox{0.67}{
  \begin{tabular}{lccccc}
    \toprule
    \textbf{Dataset} & \textbf{\#Camera} & \textbf{FPS} & \textbf{\#Scenes} & \textbf{\#Frames} & \textbf{Average length (s)}  \\
    \midrule
    Objaverse~\cite{objaverse} & \NA & \NA & 3k &  \NA & \NA\\
    NeRF-DS~\cite{yan2023nerf} & 2 & \NA & 8 & 10.2k & \NA \\
    Block-NeRF~\cite{tancik2022blocknerf} & 12 & 10 & 1 & 12k & 100\textsuperscript{*} \\
    HyperNeRF~\cite{park2021hypernerf} & 1 & 15 & 17 & 13.8k & 27 \\
    OMMO~\cite{lu2023large} & 1 & \NA & 33 & 14.7k & \NA \\
    Eyeful Tower~\cite{VRNeRF} & 22 & < 4 & 11 & 28.6k & \textbf{2k$^{\dagger}$} \\
    DyNeRF~\cite{dynerf} & 18 & 30 & 6 & 37.8k & 10 \\
    ILFV~\cite{broxton2020immersive} & 46 & 30 & 15 &  270.4k & 13 \\
    Deep3DMV~\cite{lin2021deep} & 10 & \textbf{120} & \textbf{96} & 3.8M & 33 \\
    \shortname & \textbf{53} & \textbf{120} & 54 & \textbf{17.4M} &  51  \\ 
    \bottomrule
  \end{tabular}
  }
  \caption{Specifications of our \shortname~dataset and other dynamic datasets. $*$ indicates that despite Block-NeRF consisting of a 100s-long video, it is made up of numerous transient street scenes, each with restricted view coverage. $\dagger$ indicates that although the average length of Eyeful Tower is 2000 s, the FPS is less than 4. Our \shortname~dataset is the largest visual dataset for dynamic neural fields captured at 120 FPS with an average video length of 51 s.
  }
  \label{tab:dataset_stats}
 \end{table}
\begin{figure*}[t!] 
\centering
\vspace{-5mm}
  \includegraphics[width=\textwidth]{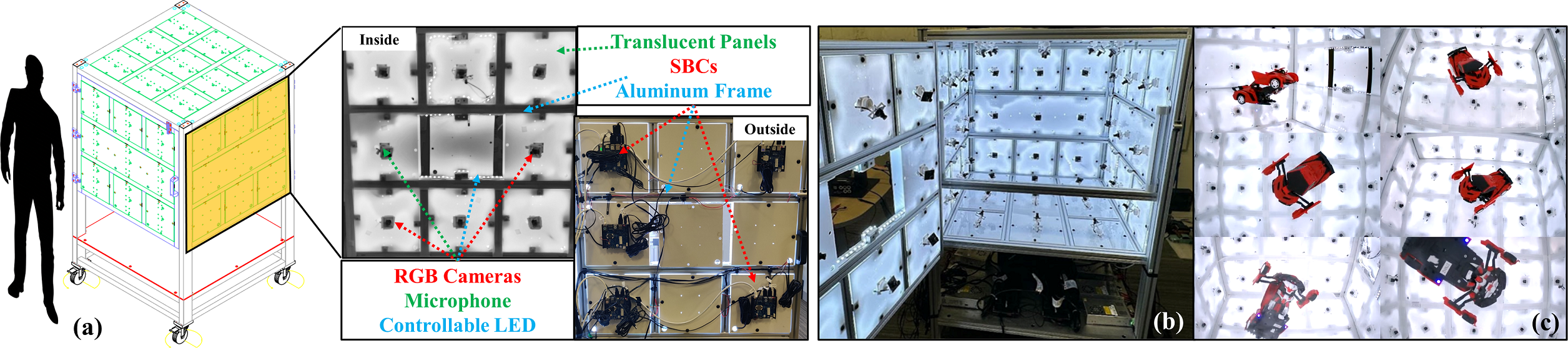}
  \vspace{-3mm}
  \caption{(a)~\brics is a refrigerator-sized aluminum frame that supports a 1~m$^3$ capture volume mounted on wheels for mobility.
  Each side wall of the capture volume is divided into a 3$\times$3 grid, with each grid square containing sensors, LEDs, single-board computers (SBCs), and light diffusers.
  (b)~Two walls of the capture volume act as doors for easy access to the capture volume.
  (c)~We can acquire 360$^\circ$ RGB views of dynamic objects and intricate hand-object interactions in this capture volume (6 views shown).
    }
  \label{fig:bricsmini}
\end{figure*}

\section{\bricsfull (\brics)}
\label{sec:hardware}
Our goal is to capture long-duration sequences of table-scale objects and interactions to enable further research in high-fidelity dynamic neural fields.
To achieve this, we need a hardware and software system that can capture high-framerate, high-resolution video and have the capability to synchronize and calibrate these sensor streams.
While commercial products exist for this purpose, they are expensive and do not meet our requirements.
We therefore designed and built our own hardware and software solution which we call the \textbf{\bricsfullu (\brics)}.
%
%
\Cref{fig:bricsmini} shows our system for capturing synchronized data.

%
%
\parahead{\brics Hardware}
Our system uses a mobile aluminum frame, housing a 1~m$^3$ capture volume outfitted with sensor panels across a 3x3 grid on each of its six sides (\Cref{fig:bricsmini}~(a)). 
These panels consist of RGB cameras, microphones, and LED light strips, which together create a versatile and uniformly light environment.
All our captures use 8-bit quantized pixels, auto exposure, and auto saturation.
For 360$^\circ$ capture, we installed a transparent shelf on which objects are placed.
The system is designed to handle large data output through a custom communication setup that compresses and transmits data to a high-capacity control workstation. 
This design, combining portability, comprehensive capture capabilities, and efficient data management, allows for dynamic, 360$^\circ$ view capturing with low latency.

\parahead{\brics Software}
While our hardware allows the capture of large-scale rich multi-view data, controlling and calibrating the cameras, and synchronizing and managing data requires specialized software.
For camera and microphone synchronization, we adopt network-based synchronization~\cite{ansari2019wireless} with an accuracy of 2--3~ms.
For camera calibration, before each capture session, we affix transparent curtains with ArUco markers to the wall and capture one calibration frame. 
We then remove the markers from the walls when capturing actual scenes.
Using COLMAP~\cite{schoenberger2016sfm,schoenberger2016mvs}, we generate camera poses for the 53 cameras.
The camera poses are further refined using I-NGP's~\cite{mueller2022instant} dense photometric loss for improved reconstruction quality.
Finally, we also built software for efficiently transferring terabytes of data from the control workstation to cloud storage.

%
%
The \shortname dynamic dataset contains synchronized long-duration videos of both moving objects and intricate hand interactions.
Our goal is to make this dataset useful for learning long-duration dynamic neural fields of appearance - existing methods~\cite{dynerf,wang2023mixed,hexplane,kplanes_2023,attal2023hyperreel,li2022streaming,nerfplayer, wu20234dgaussians, luiten2023dynamic} have been limited to only short durations, usually around 10 seconds.
Instead of just using 10~s clips of the sequences, we fully benchmark all sequences in our dynamic dataset that contains 21 object sequences categorized by different motion types, 25 hand-object interaction sequences including human daily activity, and 8 long-duration sequences with rich information.
In total, \shortname dynamic dataset contains \textbf{17.4~M} image frames of 53 dynamic scenes over ~\textbf{2738} seconds.
In addition, our data also contains masks for foreground-background segmentation.
Although not the focus of this work, we optionally provide synchronized audio and text descriptions for all sequences.
To our knowledge, this is the largest-scale dynamic dataset with a focus on table-scale interactions.
%




\section{\shortname Dataset}
\label{sec:dataset}

\begin{figure*}[t!]
  \centering
  \begin{subfigure}[b]{\textwidth}
    \includegraphics[width=\textwidth]{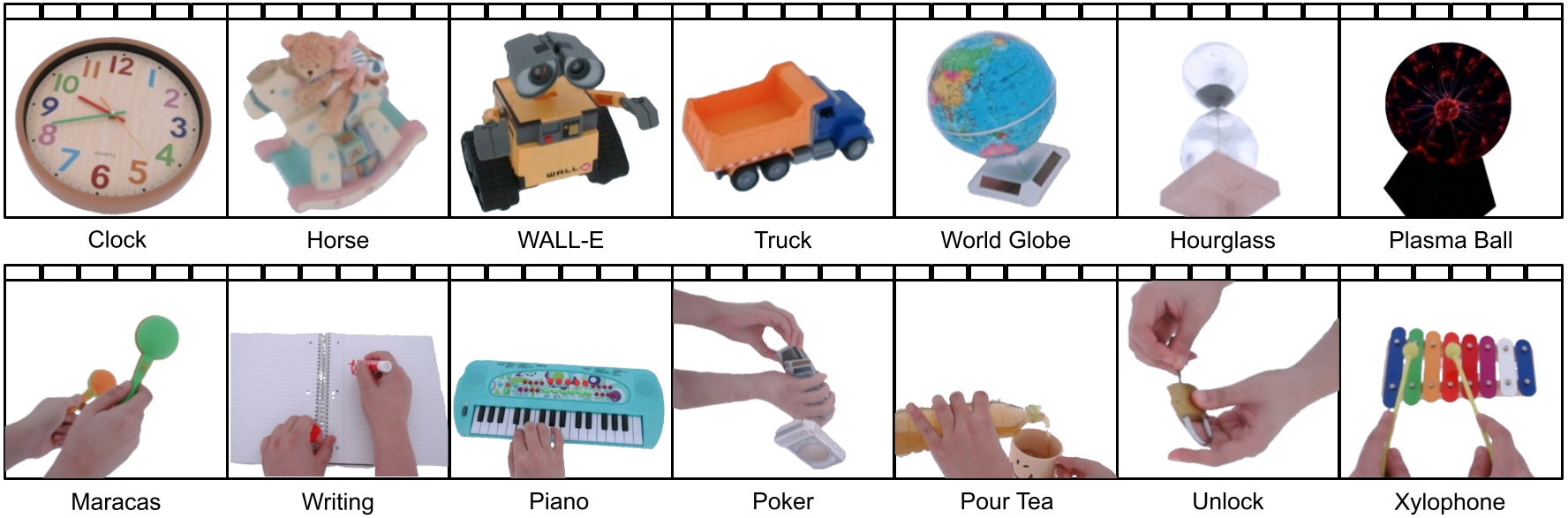}
  \end{subfigure}
  
  \caption{\shortname covers diverse object and hand-object interaction data. Our object sequences represent a variety of motion types, while our hand-object interaction data contain intricate and realistic motions.}
  \label{fig:raw_data}
\end{figure*}

\parahead{Dynamic Objects}
%
We captured 21 dynamic sequences with everyday objects and toys that move (see \Cref{fig:raw_data}).
To be representative of real-world motions, we chose objects with different types of motion (see supplementary Section 2):
(1)~Slow motion: objects that perform slow, continuous motions, \eg, music box and rotating world globe.
(2)~Fast motion: objects that move or transform drastically, \eg, remote control cars and dancing toys.
(3)~Detailed motion: objects that perform precise small motions, \eg, a clock.
(4)~Repetitive motion: objects that repeat the same motion pattern, \eg, Stirling engine and toys that sway left and right.
(5)~Random motion: objects that perform indeterministic motions, \eg, a toy that creates random patterns within a sphere.


\parahead{Interactions}
%
In addition to dynamic objects, we also include 25 hand-object interaction scenes representing intricate real-world activities (see supplementary Section 2).
The interactions included are hand activities commonly observed in everyday life (see Figure~\ref{fig:raw_data}), such as flipping a book, replacing a toy's batteries, and opening a lock.
The purpose of these hand-centric interaction data is to check whether the dynamic neural fields can generalize well to more complicated motion with occlusion from hands.
We hope these hand-centric interactions encourage future modeling of complex hand dynamics. 

\parahead{Long-Duration Sequences}
Although dynamic objects and interaction datasets have covered several long-duration videos, we further provide a long-duration dynamic dataset with 8 sequences of at least 120 seconds (see supplementary Section 2).
The existing methods have shown fast training speeds for 10s long sequences, but more efficient methods that can operate on longer sequences are needed.
Hence, this dataset is aimed at enabling future research in long-duration dynamic neural fields.


\parahead{Foreground-Background Segmentation}
%
A major challenge with neural radiance fields is the segmentation of foreground objects from background clutter.
Manually segmenting every frame is infeasible due to the quantity and view inconsistency.
Therefore, we developed a segmentation method using I-NGP~\cite{mueller2022instant}.
As preparation, we manually segment the foreground object in the first frame of one scene and train an I-NGP model on segmented images to refine coarse camera poses extracted from COLMAP~\cite{schoenberger2016sfm, schoenberger2016mvs}.
The manually segmented first frame is only used to refine camera calibration, and the refined pose is used for all downstream tasks. 
For each of the remaining frames, we fit a separate I-NGP model and progressively reduce the bounding box so that the model only renders the objects inside the bounding box but not the walls of \brics. 
These renderings are then used as segmentation masks and applied to the raw data.
To further refine the masks, we removed connected components smaller than a threshold.
Segmenting with this method is possible because all objects are placed around the center of \brics.
Since the segmentation is generated from I-NGP, the masks are multi-view consistent.

%

%
\begin{table*}[t!]
  \centering
\scalebox{1.}{
\begin{tabular}{lcccccc}
    \toprule
    \textbf{Baseline} & \textbf{PSNR}$\uparrow$ & \textbf{SSIM}$\uparrow$ & \textbf{LPIPS}$\downarrow$ & \textbf{JOD}$\uparrow$ & \textbf{Train (s/f)}$\downarrow$ & \textbf{Render (s/f)}$\downarrow$ \\
    \midrule
    PF I-NGP~\cite{mueller2022instant} & \textbf{28.31} $\pm$ 3.27 & \textbf{0.94} $\pm$ 0.03 & \textbf{0.08} $\pm$ 0.04 & \textbf{7.61} $\pm$ 0.88 & 48.70 $\pm$ 4.40 & \textbf{0.94} $\pm$ 0.25 \\ 
    MixVoxels~\cite{wang2023mixed} & 27.68 $\pm$ 2.51 & \textbf{0.94} $\pm$ 0.03 & 0.09 $\pm$ 0.04 & 7.56 $\pm$ 0.94 & 57.55 $\pm$ 6.96 & 1.48 $\pm$ 0.49\\ 
    K-Planes~\cite{kplanes_2023} & 26.39 $\pm$ 3.13 & 0.92 $\pm$ 0.03 & 0.19 $\pm$ 0.07 & 7.18 $\pm$ 1.08 & \textbf{47.59} $\pm$ 5.13 & 3.03 $\pm$ 0.20\\  
   
    \bottomrule
  \end{tabular}}
  \caption{We compare the rendering quality and train/render time of PF I-NGP, MixVoxels, and K-Planes for dynamic scenes. Surprisingly, PF I-NGP achieves higher rendering quality and equal or even faster training speed than MixVoxels and K-Planes without directly using temporal information from the adjacent frames.
  }
  \label{tab:dynamic_benchmark}
\end{table*}

\section{Benchmarks \& Experiments}
\label{sec:bench_exp}
In this section, we show how \shortname can be used to benchmark dynamic neural field methods using standardized metrics, we analyze the effect of critical parameters on these methods, and justify the need for our dataset.
These experiments were performed on Nvidia GPUs (RTX 3090, A5000) and involved \textbf{over 500 GPU-days} of training and inference.

\subsection{Benchmark Comparisons} 
\label{sec:bench}
Our goal is to compare and contrast state-of-the-art methods for dynamic neural field reconstruction on our dataset.
Specifically, we choose to compare three methods: (1)~Per-Frame I-NGP (PF I-NGP)~\cite{mueller2022instant}, a NeRF model which we train on individual frames in all 54 sequences, (2)~MixVoxels~\cite{wang2023mixed}, a state-of-the-art dynamic neural radiance field which uses variation fields to decompose scenes into static and dynamic voxels, and (3)~K-Planes~\cite{kplanes_2023} which encourages natural decomposition through planar factorization with L1 regularization for space-time decomposition.
We plan to include very recent work such as \cite{luiten2023dynamic,wu20234dgaussians} in the future.

\parahead{Pre-processing}
We downsample all our sequences to 30~FPS and then segment all frames following \Cref{sec:dataset}.
We split all of our sequences into 5-second chunks (150 frames with 30~FPS, except for PF I-NGP, which has chunk size 1) and then train the above methods per chunk.
We select 35 out of 53 best cameras for training and hold out 6 cameras for testing.
Specifically, we eliminate the cameras from the bottom row of the side panels due to reflection caused by the glass panel in \brics and randomly select one camera from each panel as test views.
We note that current neural radiance fields cannot handle reflections from the glass panel on which we place the objects for 12 views in our data.
Considering that not every NeRF model supports camera distortion parameters, we undistort the images with OpenCV~\cite{opencv_library} and crop the images to the same size ($1160\times550$) after undistortion.

\parahead{Metrics}
%
We use (a) Peak Signal-to-Noise Ratio (PSNR), (b) Structural Similarity Index Measure (SSIM)~\cite{sara2019image, journals/corr/UpchurchSB16}, and (c) Learned Perceptual Image Patch Similarity (LPIPS)~\cite{zhang2018perceptual} to measure the rendering quality, and Just Objectionable Difference (JOD)~\cite{mantiuk2021fovvideovdp} to measure the visual difference between rendered video and ground truth, along with per-frame training/rendering time (in seconds) for 6 testing views.

\parahead{Results}
We quantitatively compare the three methods in \Cref{tab:dynamic_benchmark}.
Surprisingly, although PF I-NGP is trained on each frame individually without directly utilizing temporal information, its reconstruction quality is better than both MixVoxels and K-Planes in terms of PSNR, SSIM, and LPIPS.
However, PF I-NGP suffers from temporal inconsistency, which is especially obvious for static parts (see \Cref{fig:tmp_2} and supplementary Figure~21).
Furthermore, MixVoxels only requires 2.7-4.7~MB storage space per time step, and K-Planes requires 2~MB, both of which are over six times smaller than PF I-NGP's 29~MB.
Although MixVoxels is designed for dynamic scenes, its training and inference times are higher than PF I-NGP (with a higher variance).
K-Planes has training times similar to PF I-NGP but has significantly longer inference times.
Besides, we also notice that MixVoxels struggles to capture the dynamic components of the scenes, leading to blurry and noisy reconstruction (see \Cref{fig:teaser2,fig:tmp_2}). 
We hypothesize that this is caused by insufficient capacities of the dynamic voxels when there are a lot of dynamic samples.
In contrast, K-Planes struggles to capture the static components, such as the background of the scenes, especially in the parts where there is little or no motion. 
This could be the result of overfitting and contamination from the dynamic planes due to incorrect space-time decomposition.

\subsection{Experimental Analysis} 
\label{sec:dy_exp}
The goal of this section is to identify whether the dynamic neural fields are sensitive to temporal information and spatial information. 
For the experiment, we select sequences longer than 30 seconds from the object and interaction dataset.
We then use the first 30 seconds (900 frames) of these sequences for the following experiments.

\begin{figure}[tbh!]
  \centering
  \begin{subfigure}[b]{0.48\textwidth}
    \includegraphics[width=\textwidth]{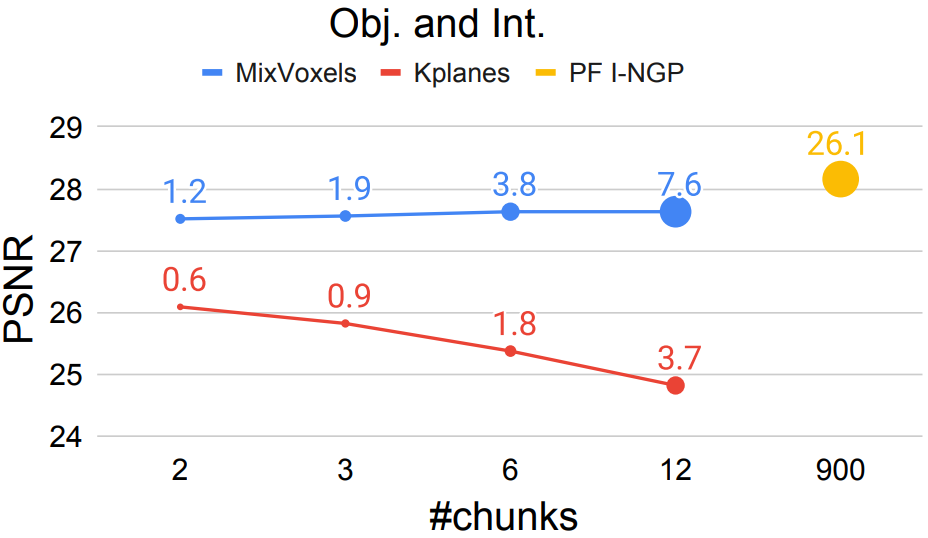}
  \end{subfigure}
  
  \caption{The rendering quality across different numbers of chunks with object and interaction data. The circle dot presents the storage space of the models in GB. MixVoxels prefers less temporal information, while K-Planes prefers more temporal information.}
  \label{fig:exp_b_chart1}
\end{figure}

\begin{figure}[tbh!]
  \centering
  \begin{subfigure}[b]{0.48\textwidth}
    \includegraphics[width=\textwidth]{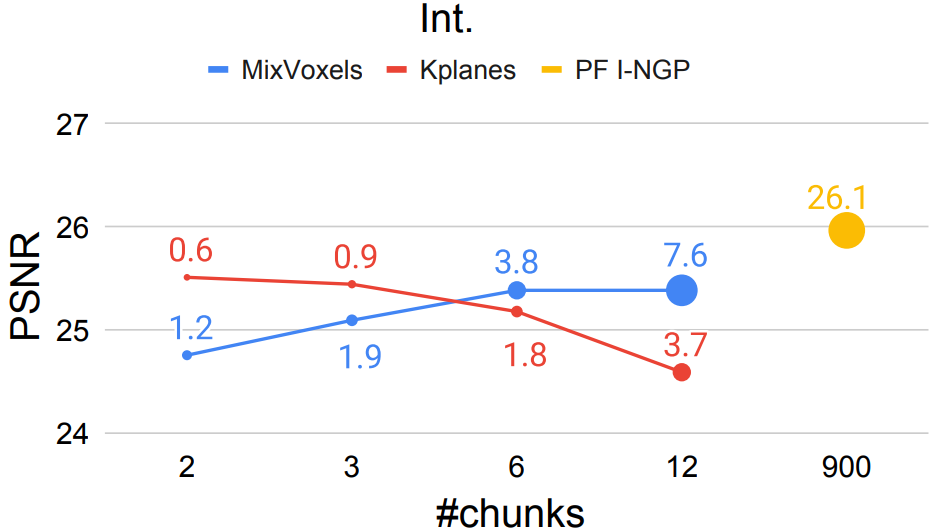}
  \end{subfigure}
  
  \caption{The rendering quality across different numbers of chunks with interaction data only. The circle dot presents the storage space of the models in GB. K-Planes outperforms MixVoxels with more temporal information on complex motion data.}
  \label{fig:exp_b_chart2}
\end{figure}

\begin{figure*}[t!]
  \centering
  \begin{subfigure}[b]{\textwidth}
    \includegraphics[width=\textwidth]{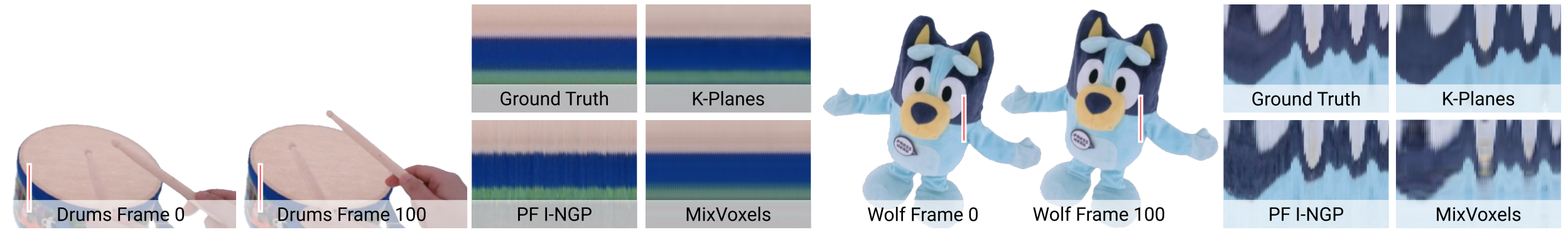}
  \end{subfigure}
  
  \caption{Visualization of temporal consistency in the same views by concatenating pixels from the same line across time steps. If a method is temporally consistent, its figure should be smooth horizontally and similar to the ground truth. PF I-NGP is less consistent across time, especially for static parts (e.g., drum), while MixVoxels is noisier on the dynamic parts (e.g., wolf's body).
  }
  \label{fig:tmp_2}
\end{figure*}

\parahead{Temporal Information}
In theory, temporal information can improve the performance of learning-based methods~\cite{hochreiter1997long, selva2023video}.
However, benchmark results in ~\Cref{sec:bench} demonstrate that PF I-NGP outperforms MixVoxels and K-Planes. 
To further investigate how sensitive these methods are to the temporal information, we split the 30-second long sequences into 2, 3, 6, and 12 chunks and train one dynamic NeRF model per chunk.
\Cref{fig:exp_b_chart1} shows that Mixvoxels performs slightly better when trained with less temporal information (more chunks), but its performance remains roughly the same across different numbers of chunks. 
The dynamic branch of MixVoxels may not have sufficient capacity to handle more dynamic samples. 
Unlike MixVoxels, K-Planes is more sensitive to sequence lengths and performs better with more temporal information.
Through the rendering results, we found that fewer chunks also mitigate the overfitting problem of K-Planes on \shortname (see supplementary Figure 12). 
One interesting finding is that although MixVoxels outperforms K-Planes with object and interaction data, K-Planes outperforms MixVoxels with 2 and 3 chunks setting on interaction data (see Figure~\ref{fig:exp_b_chart2}). 
This indicates that K-Planes can better handle the more complex motions in the interaction data when provided with more temporal information.



\parahead{Spatial Information}
Intuitively, neural fields trained with higher-resolution images should result in better reconstruction quality.
To test this hypothesis, we conduct comparisons of model performances across different resolutions.
In this experiment, we train the three methods on lower resolutions by downscaling the training set from $1160\times550$ (undistorted images) to $674\times320$ and $464\times220$. 
After training, we evaluate the trained models by rendering test views at the original resolution (see \Cref {tab:exp_h} and supplementary Figure 13). 
To our surprise, we found that the performance of PF I-NGP remains almost the same, and rendering results have apparent fine-grained details.
Furthermore, MixVoxels and K-Planes perform better at lower resolution, with MixVoxels performing the best at $674\times320$ and K-Planes performing the best at $464\times220$. 
Both methods suffer from similarly blurry details across all resolutions.
This interesting result contradicts our hypothesis, and there could be several reasons for it.
First, NeRFs have shown impressive spatial interpolation ability leading to only a minimal drop in performance with reducing resolution.
Second, under the same training setting, NeRFs will revisit the training samples more frequently when trained on lower-resolution images and thus reconstruct these samples better. 
Finally, dynamic NeRFs need to spend much more of their capacities to capture moving objects which could result in insufficient capacity to capture fine-grained details.
In conclusion, we suspect that current dynamic methods cannot efficiently utilize spatial information in high-resolution images as they are biased toward motion and misses the high-frequency details presented in the images.
In addition, we note that human perception of these images may not match the observed quantitative results~\cite{blau2018perception}.

\begin{table}[tbh!]
   \centering
   \scalebox{0.9}{
   \begin{tabular}{lccc}
     \toprule
     \textbf{Baseline} & \textbf{$\mathbf{1160\times550}$} & \textbf{$\mathbf{674\times320}$} & \textbf{$\mathbf{464\times220}$} \\
     \midrule
      PF I-NGP~\cite{mueller2022instant} & 28.16 & 28.19 & 28.15 \\
      MixVoxels~\cite{wang2023mixed} & 27.63 & 27.75 & 26.91 \\
      K-Planes~\cite{kplanes_2023} & 25.38 & 25.57 & 26.03 \\
     \bottomrule
   \end{tabular}
   }
   \caption{The PSNR of each baseline across different resolutions. During testing, we interpolate the images to $1160\times550$ resolution. The performance of PF I-NGP remains similar. MixVoxels and K-Planes get slightly better performance with the low-resolution training set.}
     \label{tab:exp_h}
\end{table}

\parahead{Spatial and Temporal Information}
In the previous experiment, we only have one control variable, spatial resolution or temporal length. 
It is unclear whether the same conclusion will hold if we change spatial resolution and temporal length simultaneously. 
Hence, in this experiment, we change spatial resolution and temporal length simultaneously while maintaining a similar size of the 3D volume (width x height x temporal length). 
Table~\ref{tab:exp_c} shows MixVoxels performs worst, and K-Planes performs best with the lowest spatial resolution and longer temporal length setting ($464\times220$, 900 frames). 
This matches our findings in the previous experiments. 
If we check Table~\ref{tab:exp_h} and Table~\ref{tab:exp_c} together, the phenomenon is more obvious.
The PSNR of MixVoxels drops from 27.75 to 27.2 and 26.91 to 26.1 for spatial resolution $674\times320$ and $464\times220$, respectively, after including more temporal information.
The PSNR of K-Planes increases from 25.57 to 26.04 and 26.03 to 26.19 for spatial resolution $674\times320$ and $464\times220$, respectively, after including more temporal information.

\begin{table}[tbh!]
   \centering
   \scalebox{0.7}{
   \begin{tabular}{lccc}
     \toprule
     \textbf{Baseline} & \textbf{$\mathbf{1160\times550}$, 6 ch.} & \textbf{$\mathbf{674\times320}$, 2 ch.} & \textbf{$\mathbf{464\times220}$, 1 ch.} \\
     \midrule
      MixVoxels~\cite{wang2023mixed} & \textbf{27.63}  & \textbf{27.20} & 26.10 \\
      K-Planes~\cite{kplanes_2023} & 25.38 & 26.04 & \textbf{26.19} \\
     \bottomrule
   \end{tabular}
   }
   \caption{The PSNR of each baseline across different spatial resolutions and temporal lengths. MixVoxels reaches the worst performance, and K-Planes reaches the best performance with the lowest spatial resolution and fewest chunks setting ($464\times220$, 1 chunk).}
     \label{tab:exp_c}
\end{table}

\subsection{Dataset Justification}
In this section, we justify the need for our 360$^\circ$ views with 53 cameras, and other design choices.

\parahead{Number of Cameras}
For evaluating the number of cameras, we compare three settings: (1) \textit{All-view}, which follows the original setting with all cameras, (2) \textit{Forward}, which only uses the cameras from two adjacent side panels, resulting in 10 cameras, (3) \textit{Multi-view}, which uses two cameras per side panel and one camera from top and bottom panel, resulting in 10 cameras. 

%
Both quantitative and qualitative results (see \Cref{tab:exp_g} and supplementary Figure 15-16) demonstrate that \textit{All-view} outperforms \textit{Multi-view}, indicating that more cameras improve the rendering quality of NeRFs. 
In addition, \textit{All-view} and \textit{Multi-view} outperform \textit{Forward} when tested on an occluded view, suggesting that multi-view 360$^\circ$ is better than forward-facing settings for benchmarking.

\begin{table}[tbh!]
   \centering
   \scalebox{0.85}{
   \begin{tabular}{lccc}
     \toprule
     \textbf{Baseline} & \textbf{All-view} & \textbf{Forward} & \textbf{Multi-view} \\
     \midrule
      PF I-NGP~\cite{mueller2022instant} & 28.16 / 27.24 & 24.77 / 23.99 & 24.07 / 25.37 \\
      MixVoxels~\cite{wang2023mixed} & 27.63 / 27.43 & 20.51 / 15.80 & 23.65 / 24.02 \\
      K-Planes~\cite{kplanes_2023} & 25.38 / 24.55 & 23.23 / 22.40 & 22.64 / 22.89 \\
     \bottomrule
   \end{tabular}
   }
    \caption{The PSNR of testing views / occluded views across different settings of the capture system. The PSNR of \textit{All-view} is higher than \textit{Multi-view}. Hence, more cameras can help NeRFs. The PSNRs of \textit{All-view} and \textit{Multi-view} are higher than \textit{Forward} on occluded view, indicating that multi-view 360$^\circ$ settings are better than forward-facing settings.}
    \label{tab:exp_g}
\end{table}

\parahead{Foreground-Background Segmentation Method}
In \Cref{sec:dataset}, we mention that we use I-NGP to segment each frame. 
Although the segmentation model can be replaced with improved models in the future, we believe that our current method is suitable for \shortname, especially due to its multiview consistency. 
To validate the performance of I-NGP segmentation, we compare it against Segment Anything (SAM)~\cite{kirillov2023segany} in terms of segmentation quality and multiview consistency.
For this benchmark, we manually segment one frame of 6 random views from all scenes as ground truth and compute mean intersection over union (mIoU) for images segmented by SAM and our method.

According to \Cref{tab:exp_seg}, I-NGP segmentation reaches better mIoU and lower average standard deviation over six views on \shortname. 
In addition, the visualization results (see supplementary Figure 18 and 19) also support the statement that the performance of I-NGP segmentation is more multiview consistent.

\begin{table}[tbh!]
   \centering
   \scalebox{0.85}{
   \begin{tabular}{lccc}
     \toprule
     \textbf{Baseline} & \textbf{Obj. and Int.} & \textbf{Obj.} & \textbf{Int.} \\
     \midrule
      I-NGP Seg.~\cite{mueller2022instant} & 0.926 / 0.048 & 0.962 / 0.016 & 0.901 / 0.071 \\
      SAM~\cite{kirillov2023segany} & 0.919 / 0.086 & 0.955 / 0.042 & 0.885 / 0.118 \\
     \bottomrule
   \end{tabular}
   }
    \caption{The mean intersection over union (mIoU) / average standard deviation of mIoU over six views. I-NGP segmentations outperform SAM on \shortname. In addition, a lower standard deviation indicates more equal quality across views.}
    \label{tab:exp_seg}
\end{table}

\section{Conclusion}
\label{sec:conclusion}
We have introduced \shortname, a real-world 360$^\circ$ \underline{d}ynam\underline{i}c \underline{v}isu\underline{a}l dataset that contains synchronized long-duration sequences of table-scale moving objects and interactive scenes.
We propose a new \brics capture system for synchronized long-duration data capture, which also acts as a rich multimodal data capturing system (see supplementary Section 1).
\shortname consists of a dynamic dataset of high-resolution, high-framerate, long (5s to 3~mins), and synchronized videos captured simultaneously from 53 RGB cameras within the capture space. In total, \shortname contains 17.4~M images.

We benchmark the existing state-of-the-art dynamic neural fields with \shortname dynamic dataset and demonstrate that there is still room for improvement in terms of training and rendering speed, hardware requirement, imbalance capacity, temporal information, and spatial information.

\parahead{Limitations and Future Work}
Although \brics can also act as a multimodal capturing system, our current metrics and evaluation are limited to images  -- in future work, we will consider metrics for audio and text (see supplementary Section 7).
\brics cannot capture scenes larger than table-scale -- we plan to expand the capture system to larger volumes in the future.
Due to the training speed of the existing state-of-the-art methods, we cannot include more baselines or metrics for longer videos.
Hence, we hope to include more long sequences and baselines in the future~\cite{wu20234dgaussians, xu20234k4d}.


\parahead{Societal/Ethical Impact}
Our dataset does not reveal any private information and presents limited means for misuse. However, future extensions of our work could contain private information that can be misused.
Another possible impact is the environmental cost since the total GPU running days of training, rendering, and experiments are at least 500 days.
Thus, we will release pretrained weights. 


\section{Acknowledgements}
This work was supported by NSF grants CAREER \#2143576 and CNS \#2038897, ONR grant N00014-22-1-259, ONR DURIP grant N00014-23-1-2804, a gift from Meta Reality Labs, an AWS Cloud Credits award, and NSF CloudBank.
Arnab Dey was supported by H2020 COFUND program BoostUrCareer under Marie SklodowskaCurie grant agreement \#847581.
We thank George Konidaris, Stefanie Tellex, Rohith Agaram, and the Brown IVL team.
{
    \small
    \bibliographystyle{ieeenat_fullname}
    \bibliography{main}
}


\clearpage



\section*{Supplementary Material}

\setcounter{section}{0}

\section{Design of Brown Interaction Capture System (BRICS)}
%


\parahead{Aluminum Frame}
To capture table-scale scenes, we chose a refrigerator-sized aluminum frame (Figure 3~(a) of the main paper) that houses a 1~m$^3$ capture volume mounted on wheels for mobility.
Each of the 6 side walls of the capture volume is composed of a 3$\times$3 grid with dual polycarbonate panels on each grid square (total of 54 squares).
Two of the walls are doors that allow quick access to the capture volume.
The height of the system allows an average person to easily reach into the volume for interaction capture.
A transparent polycarbonate shelf in the capture volume allows bottom cameras to still see objects to provide a 360$^\circ$ view.
A shelf in the bottom houses power supplies, network switches, and a control workstation.

\parahead{Sensor/Illumination Panels}
For 53 of the 54 grid squares (we leave one out for easy access) on the side walls, we installed translucent polycarbonate panels on the interior consisting of cameras, microphones, and LEDs.
Each panel can support up to 3 RGB cameras, 3 microphones, and a fully programmable RGB light strip with 72 individual LEDs.
This panel naturally diffuses the LED lights enabling uniform lighting of the volume.
In our current setup, each of the 53 panels has an LED strip and 1 off-the-shelf RGB camera capturing at 1280$\times$720 @ 120~FPS.
We install microphones on 6 panels, one on each side wall of the capture cube.

\parahead{Communication Panels}
The sensor panels collectively generate more than 13.25~GB/s (0.25~GB/s per panel) of uncompressed data -- well beyond the bandwidth of common wired communication technologies like USB or ethernet.
To enable the capture and storage of such amounts of data, we built our own communication system.
Briefly, this system consists of single-board computers (SBCs) that connect to sensors via USB and are responsible for compressing the data before sending it to a control station over gigabit ethernet.
With this setup, we are able to simultaneously transmit large amounts of data with low latency.

\parahead{Control Workstation}
We use a workstation with 52 CPU cores to simultaneously uncompress, store, and transmit all the data.
To ensure high throughput, we use a 10~Gigabit ethernet uplink to the SBCs, a PCI solid state drive, and 200~GB RAM for caching.

\parahead{Panels}
\brics panels are designed to be modular, to allow for quick customization for different research endeavors. \brics consists of 42 panels in total across six sides. Each side has six square panels of size (9.75~in x 9.75~in) and a single rectangular middle panel of size (32.25~in x 9.75~in) that can be changed to consist of three square panels based on research tasks. The panels inside are white translucent panels made of TAP plastics Satinice White Acrylic to encourage light dispersion towards the inside. Outer panels are white and opaque made of TAP plastics KOMATEX foamed PVC Sheets.

The inner panels allow the mounting of three different cameras or other accessories. Although the panel currently consists of an RGB camera of size (71.5~mm x 71.5~mm) mounted at the center, future plans include attaching depth and infrared cameras. All panels are 1/8th inch in thickness. 

\parahead{Mounts}
We utilize custom-designed mounts to attach cameras to the panels. We use custom-designed ball bearing mounts, that are rotatable to allow for changing the camera orientation.

\parahead{Lighting}
We use BTF-Lighting WS2812-B individually addressable RGB lighting strips. This allows for highly customized lighting conditions and environment maps. Moreover, using LED strips allows us to add additional lighting quickly. 

Each panel has 70 LED's placed in between the inner panel and outer panel. These LED's are powered individually and sequentially connected for data. The LED's are all controlled with six Raspberry Pi 3 Model B+ computers, one for each side. To control the LED's we used the standard NeoPixel python library. Furthermore, each side allows for individual brightness control.

\parahead{Cameras}
We used off-the-shelf USB 2.0 cameras that can capture 1280$\times$720 @ 120~FPS.
Specifically, we used the ELP-SUSB1080P01-LC1100 from ELP Cameras.

\parahead{Single Board Computers (SBCs)}
We need the single board computers to have enough processing power and USB ports to support up to 3 cameras and 1 microphone each.
For this reason and easy market availability, we chose the Odroid N2+ 4~GB, which was sufficient for our purpose.

\section{Dynamic Dataset Benchmark Comparison}

\parahead{Pre-processing}
To do a benchmark, we pre-process the raw data captured through \brics following Section 5. 

\begin{table*}[tbh!]
  \centering
  \begin{tabular}{lccccc}
    \toprule
    Types & Baseline & PSNR$\uparrow$ & SSIM$\uparrow$ & LPIPS$\downarrow$ & JOD$\uparrow$ \\
    \midrule
    \multirow{3}{*}{Dynamic objects} & PF I-NGP~\cite{mueller2022instant} & 29.62 / 4.42 & 0.95 / 0.02 & 0.06 / 0.02 & 7.53 / 1.28 \\
    & MixVoxels~\cite{wang2023mixed} & 28.43 / 2.87 & 0.95 / 
    0.02 & 0.06 / 0.03 & 7.48 / 1.31\\ 
    & K-Planes~\cite{kplanes_2023} & 26.62 / 4.41 & 0.91 / 0.04 & 0.21 / 0.09 & 6.66 / 1.47 \\
    \midrule
    \multirow{3}{*}{Interactions} & PF I-NGP & 26.89 / 1.76 & 0.94 / 0.03 & 0.09 / 0.04 & 7.68 / 0.51 \\
    & MixVoxels & 26.45 / 1.91 & 0.93 / 
0.03 & 0.10 / 0.04 & 7.50 / 0.70 \\ 
    & K-Planes & 26.18 / 2.08 & 0.93 / 0.03 & 0.17 / 0.06 & 7.55 / 0.49 \\ 
    \midrule
    \multirow{3}{*}{Long-duration Sequences} & PF I-NGP & 29.35 / 0.90 & 0.93 / 0.03 & 0.10 / 0.03 & 7.60 / 0.57 \\
    & MixVoxels & 29.62 / 0.75 & 0.93 / 0.02 & 0.11 / 0.02 & 7.87 / 0.29 \\
    & K-Planes & 26.45 / 1.86 & 0.92 / 0.02 & 0.21 / 0.04 & 7.27 / 0.68 \\
    \bottomrule\\
  \end{tabular}
  \caption{Rendering quality (mean / standard deviation) of dynamic objects, interactions, and long duration sequences respectively using PF I-NGP\cite{mueller2022instant}, MixVoxels.\cite{wang2023mixed}, and K-Planes~\cite{kplanes_2023}. Although the PSNR of PF I-NGP is much better than MixVoxels in terms of dynamic objects, the PSNR of PF I-NGP is similar to the MixVoxels for interactions and even underperforms MixVoxels for long-duration sequences. This indicates that it is hard to capture the scene occluded by hands and maintain the temporal consistency, especially for static parts such as the chessboard, without utilizing the temporal information. Unlike PF I-NGP and MixVoxels, the performance of K-Planes is more consistent across the three types.}
    \label{tab:dynamic_benchmark_object_interaction}
\end{table*}

\parahead{Baselines Training}
Per-Frame I-NGP (PF I-NGP) sequentially learns a model for each time step. 
Each I-NGP~\cite{mueller2022instant} is initialized from the model of the previous time step. 
The PF I-NGP allows us to fit the dynamic video efficiently while not considering the motion between frames. 
In addition, the streamable training feature also allows us to optimize the camera pose and lens distortion individually for each frame. 
We train each I-NGP for 5000 iterations. 
The average training time for each I-NGP is 48.7 seconds with a standard deviation of 4.4 seconds. 
The smaller standard deviation is due to the fact that PF I-NGP does not consider motion.
However, this will lead to temporal inconsistency, which is especially obvious at static parts when comparing PF I-NGP with MixVoxels~\cite{wang2023mixed} and K-Planes~\cite{kplanes_2023}. 
In Figure~\ref{fig:tmp_inc}, we concatenate the pixels from the line across frames in the same view and demonstrate that PF I-NGP contains much white noise and is less temporal smooth.

MixVoxels~\cite{wang2023mixed} is trained to capture the dynamic video every 150 frames. 
We train each MixVoxels for 25000 iterations. 
We lower the dynamic threshold to capture more dynamic samples for scenes with drastic motion. 
Although MixVoxels is trained and benchmarked with black backgrounds, we render it with white backgrounds for all figures in the paper.
This will make the floaters of MixVoxels darker in the figures.
Unlike PF I-NGP, MixVoxels allows us to learn a dynamic NeRF with motion.
In addition, MixVoxels trained with multiple frames also encourage temporal consistency (see Figure~\ref{fig:tmp_inc}), which is absent in PF I-NGP. 
However, MixVoxels struggles to capture dynamic parts with complex motion and small dynamic objects.
Hence, in Figure~\ref{fig:fail_exp}, we can observe many floaters and noise around the hand of Chess, and cannot see the small train in the Music Box and holes of knitted fabric in the Crochet.
We assume the noise and floaters are caused by the insufficient capacity of the dynamic branch when receiving too many dynamic samples.
The small objects and fine-grained details are missed because the variation field fails to decompose the scene correctly.
For each time step, the average training time is 57.55 seconds, with a standard deviation of 6.96 seconds. The large standard deviation is because MixVoxels will sample more dynamic points for its dynamic branch when learning a scene with complex motion.

For a fair comparison, we also train a K-Planes every 150 frames. 
We train each K-Planes for 90000 iterations. 
Similar to the MixVoxels, K-Planes is also more temporally consistent. 
This is especially obvious in Chess and Put Fruit of Figure~\ref{fig:tmp_inc}. 
K-Planes sometimes fails to reconstruct static parts and tends to generate many floaters, especially for objects with slow motion such as Music Box. 
Additionally, it misses the fine-grained details of the knitted crochet in Figure~\ref{fig:fail_exp}.
This could be the result of overfitting and contamination from the dynamic planes due to weaker decomposition.
For each time step, the average training time is 47.59 seconds with a standard deviation of 5.13 seconds.
Although the training time of K-Planes is faster than PF I-NGP and MixVoxels without considering standard deviation, K-Planes shows the slowest rendering speed among them. 

\begin{table*}[tbh!]
  \centering
  \begin{tabular}{lccccc}
    \toprule
    Baseline & Motion & PSNR$\uparrow$ & SSIM$\uparrow$ & LPIPS$\downarrow$ & JOD$\uparrow$ \\
    \midrule
    \multirow{5}{*}{PF I-NGP~\cite{mueller2022instant}} & Slow & 29.76 / 2.18 & 0.96 / 0.02 & 0.05 / 0.03 & 7.13 / 2.57\\
    & Fast & 30.06 / 5.15 & 0.95 / 0.01  & 0.06 / 0.02 &  7.60 / 0.60 \\
    & Detailed & 33.11 / 7.81 & 0.96 / 0.02  & 0.07 / 0.03 & 6.74 / 2.35  \\
    & Repetitive & 28.76 / 2.16 & 0.96 / 0.02  & 0.06 / 0.02 & 7.75 / 0.71 \\
    & Random & 31.59 / 8.95 & 0.95 / 0.01 & 0.07 / 0.02 & 6.74 / 2.34 \\
    \midrule
    \multirow{5}{*}{MixVoxels~\cite{wang2023mixed}} & Slow & 29.64 / 2.14  & 0.96 / 0.02 & 0.05 / 0.03 & 7.58 / 2.52 \\
    & Fast & 28.39 / 3.09 & 0.95 / 0.02  & 0.07 / 0.02 & 7.41 / 0.73 \\
    & Detailed & 30.25 / 3.95 & 0.96 / 0.02 & 0.06 / 0.03 &  6.59 / 2.57 \\
    & Repetitive & 28.37 / 2.35 & 0.94 / 0.02 & 0.09 / 0.02 & 6.94 / 0.72 \\
    & Random$^{*}$ & 27.89 / 5.75 & 0.95 / 0.02 & 0.07 / 0.03 &  5.91 / 1.99 \\
    \midrule
    \multirow{5}{*}{K-Planes~\cite{kplanes_2023}} 
    & Slow & 25.18 / 2.05 & 0.88 / 0.03 &0.31 / 0.03 &6.10 / 1.03 \\
    & Fast & 26.99 / 5.23 & 0.92 / 0.05 & 0.18 / 0.09 & 6.73 / 1.63 \\
    & Detailed & 27.91 / 8.54 & 0.88 / 0.05 & 0.26 / 0.10 & 5.68 / 1.54 \\
    & Repetitive & 25.43 / 2.78 & 0.91 / 0.05 & 0.23 / 0.10 & 6.47 / 1.60 \\
    & Random & 30.86 / 6.02 & 0.92 / 0.04 & 0.19 / 0.07 & 6.65 / 1.38 \\
    \bottomrule\\
  \end{tabular}
  \caption{Rendering quality (mean / standard deviation) of dynamic objects using PF I-NGP\cite{mueller2022instant}, MixVoxels\cite{wang2023mixed} and K-Planes\cite{kplanes_2023} in terms of different types of motion. * indicates that the results do not include the Plasma Ball Clip sequence because the codebase cannot handle that scene. PF I-NGP serves as a baseline without considering temporal information. Compared with PF I-NGP, the PSNR of MixVoxels decreases by 0.12 dB, 1.67 dB, 2.86 dB, 0.39 dB, and 3.7 dB for slow, fast, detailed, repetitive, and random motion, respectively. MixVoxels perform better on scenes with slow and repetitive motion while worse on scenes with drastic motion. Compared with PF I-NGP, the PSNR of K-Planes decreases by 4.58 dB, 3.07 dB, 5.2 dB, 3.33 dB, and 0.73 dB for slow, fast, detailed, repetitive, and random motion, respectively. K-Planes perform better on scenes with fast and random motion while worse on scenes with less drastic motion.}
    \label{tab:dynamic_benchmark_motion}
\end{table*}

\Cref{tab:dynamic_benchmark_object_interaction} separately shows quantitative results of dynamic objects, interactions, and long-duration sequences. 
PF I-NGP, MixVoxels, and K-Planes perform better with dynamic objects than interactions.
One reason is that models trained on dynamic objects do not need to handle occlusion caused by the hands. 
Another reason is that hand motion is more complex.
The PSNR gap of PF I-NGP is 2.73 dB, the PSNR gap of MixVoxels is 1.98 dB, and the PSNR gap of K-Planes is 0.44 dB. 
The performance gap between dynamic objects and interactions is more obvious with PF I-NGP because it does not utilize temporal information and, therefore, cannot handle occlusion well. 
A similar situation happens to PF I-NGP when comparing dynamic objects with long-duration sequences.
However, this situation does not occur with MixVoxels when comparing dynamic objects with long-duration sequences, which also capture hand-object interaction scenes but longer.
This is because the scenes of long-duration sequences often contain a large portion of static parts, such as the chessboard in the Chess Long (see \Cref{fig:long_chess}). 
MixVoxels can largely benefit from this kind of sequence because the independent static branch of the MixVoxels can handle static parts smoothly across frames.
Interestingly, this allows MixVoxels to outperform PF I-NGP, which is less temporally smooth across these frames (see \Cref{fig:tmp_inc}). 
Unlike PF I-NGP and MixVoxels, the performance of K-Planes is robust across three types because of the relatively soft static dynamic decomposition.

\parahead{Dynamic Object Results}
\Cref{tab:dynamic_benchmark_motion} shows the performance of PF I-NGP, MixVoxels, and K-Planes in different motion types. 
We manually classified the 21 dynamic object sequences into five overlapping categories: slow, fast, detailed, repetitive, and random (\Cref{tab:dynamic_object_types}).
Although PF I-NGP does not consider motion, it serves as a baseline for dynamic models by fitting each frame separately. 

\begin{figure*}[t!]
  \centering
  \begin{subfigure}[b]{0.82\textwidth}
    \includegraphics[width=\textwidth]{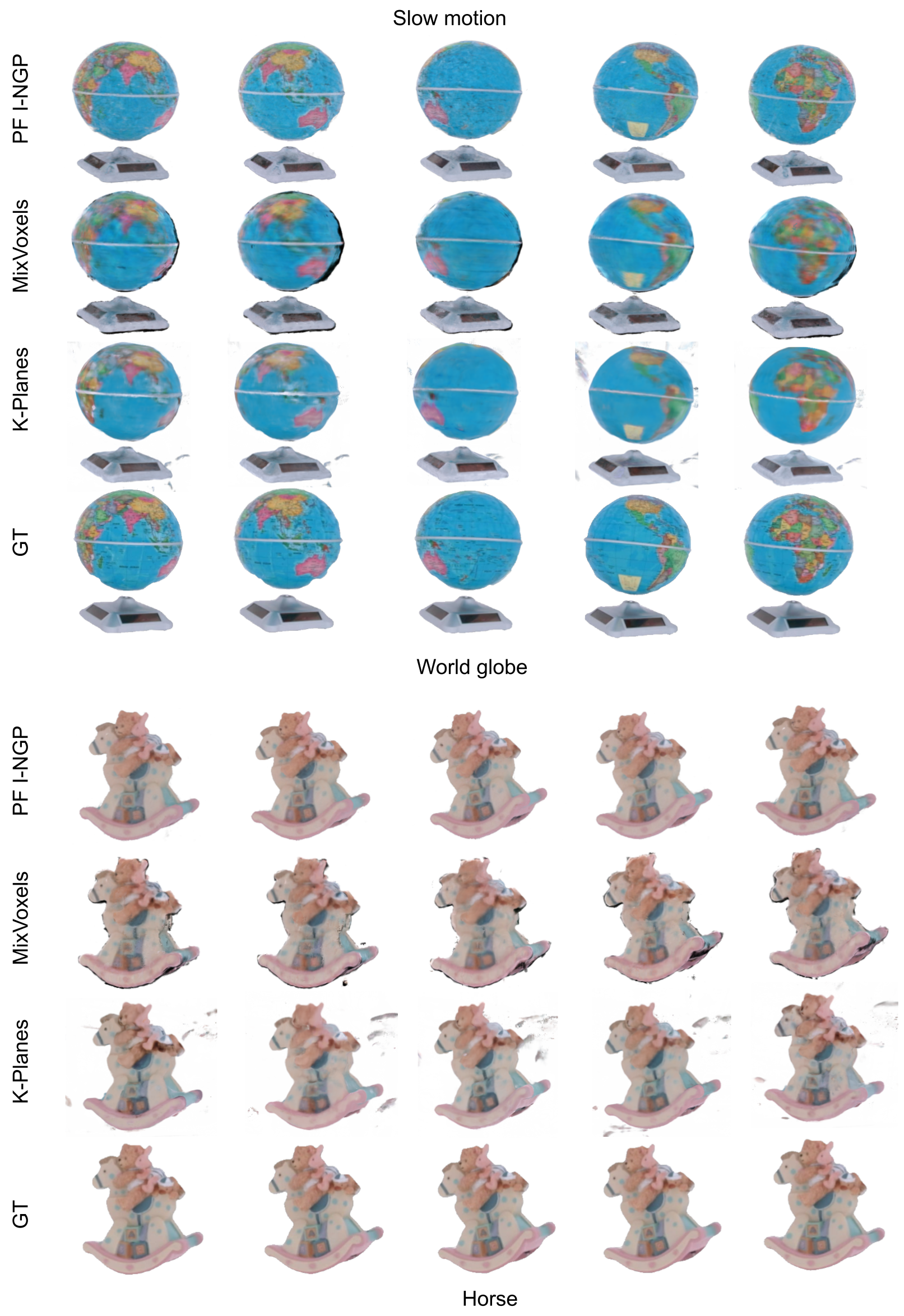}
  \end{subfigure}
  
  \caption{Test view reconstruction and ground truth of the two dynamic objects with slow motion: world globe and horse. Top: Although MixVoxels cannot capture the high-frequency details, the object's shape is correctly reconstructed. K-Planes cannot capture the high-frequency details and construct many floaters around the object. Bottom: The rendering results of MixVoxels are close to the ground truth. K-Planes still suffers from a bunch of floaters.  }
  \label{fig:slow}
\end{figure*}

\begin{figure*}[t!]
  \centering
  \begin{subfigure}[b]{0.85\textwidth}
    \includegraphics[width=\textwidth]{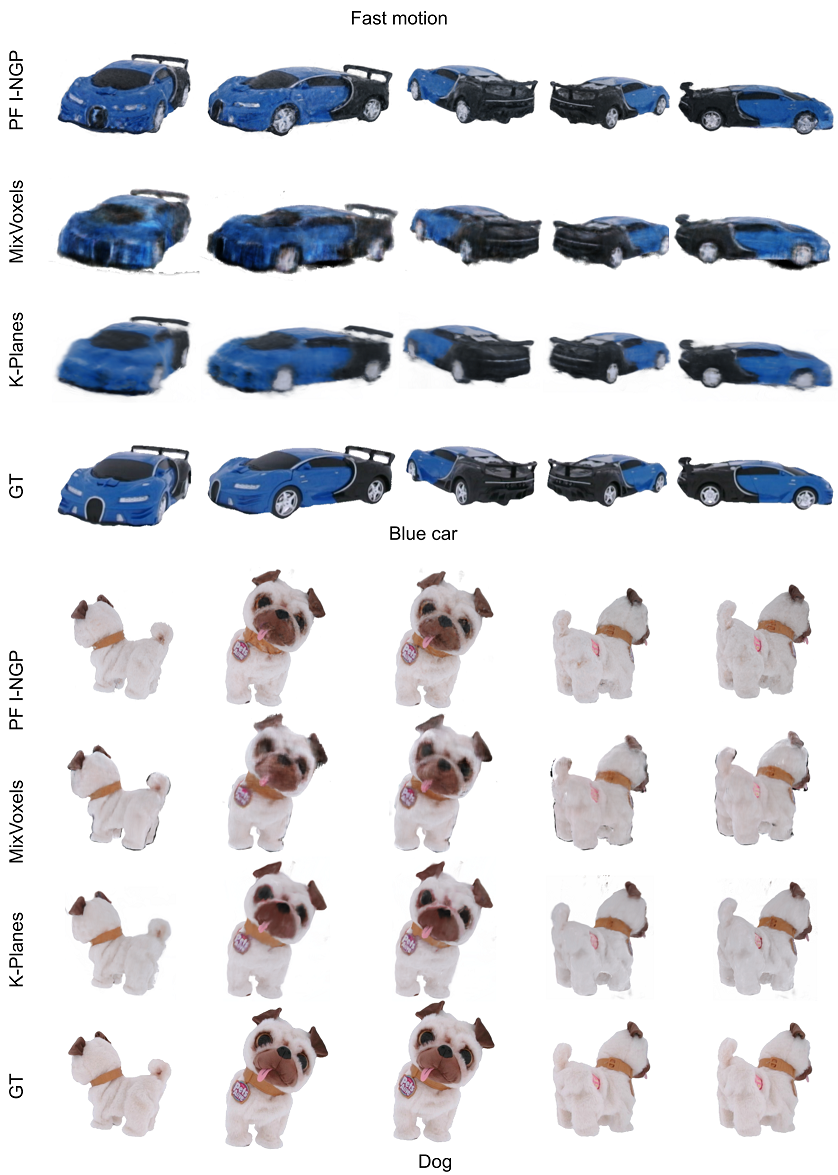}
  \end{subfigure}
  
  \caption{Test view reconstruction and ground truth of two fast motion objects: blue car and dog. Top: The reconstruction results of MixVoxels and K-Planes are blurry. Bottom: The reconstruction results of MixVoxels, such as the dog's tongue, are slightly blurry. K-Planes looks similar to the ground truth.}
  \label{fig:fast}
\end{figure*}

\begin{figure*}[t!]
  \centering
  \begin{subfigure}[b]{0.8\textwidth}
    \includegraphics[width=\textwidth]{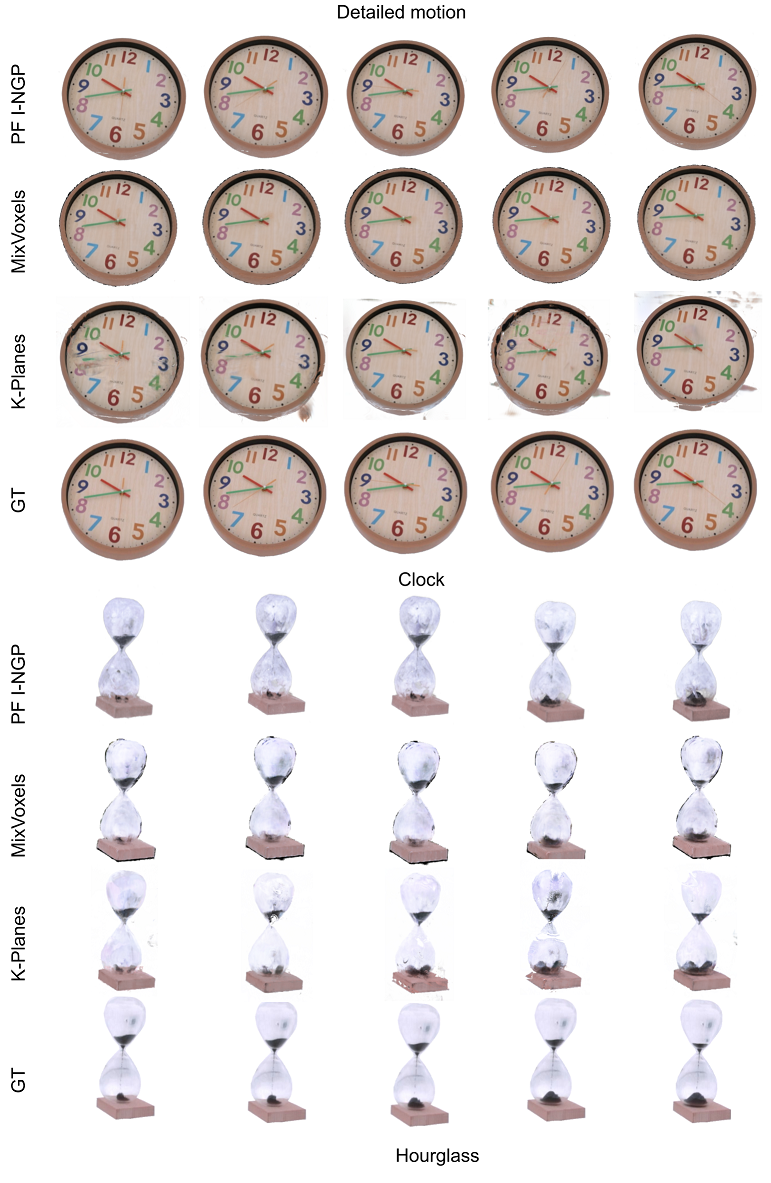}
  \end{subfigure}
  
  \caption{Test view reconstruction and ground truth of two detailed motion objects: clock and hourglass. Top: MixVoxels captures the clock's body accurately but cannot reconstruct the second hand. By contrast, K-Planes struggle for the static part, the clock's body, but the second hand is partially reconstructed. Bottom: All baselines cannot reconstruct the hourglass well due to transparency and highly detailed motion. }
  \label{fig:detailed}
\end{figure*}

\begin{figure*}[t!]
  \centering
  \begin{subfigure}[b]{0.75\textwidth}
    \includegraphics[width=\textwidth]{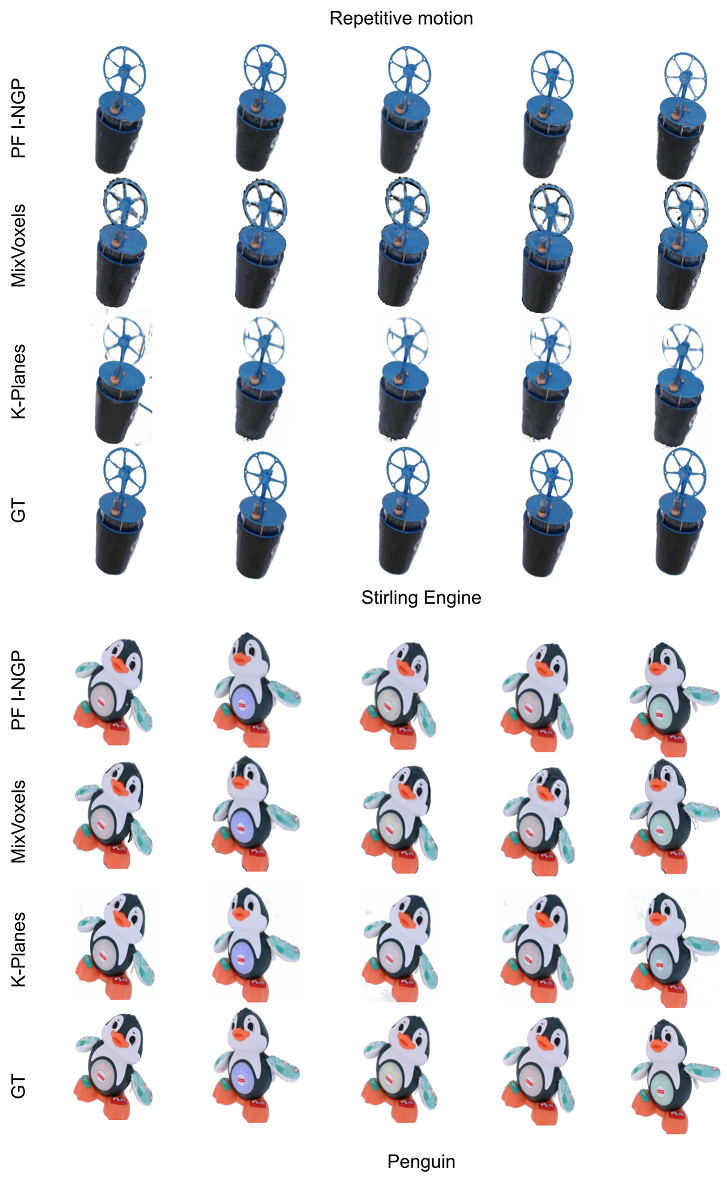}
  \end{subfigure}
  
  \caption{Test view reconstruction and ground truth of two objects with repetitive motions: stirling engine and toy penguin. Top: MixVoxels reconstructs the stirling engine well, while K-Planes fails to capture the rotating part correctly. Bottom: All baselines almost faithfully capture the toy penguin.}
  \label{fig:repetitive}
\end{figure*}

\begin{figure*}[t!]
  \centering
  \begin{subfigure}[b]{0.75\textwidth}
    \includegraphics[width=\textwidth]{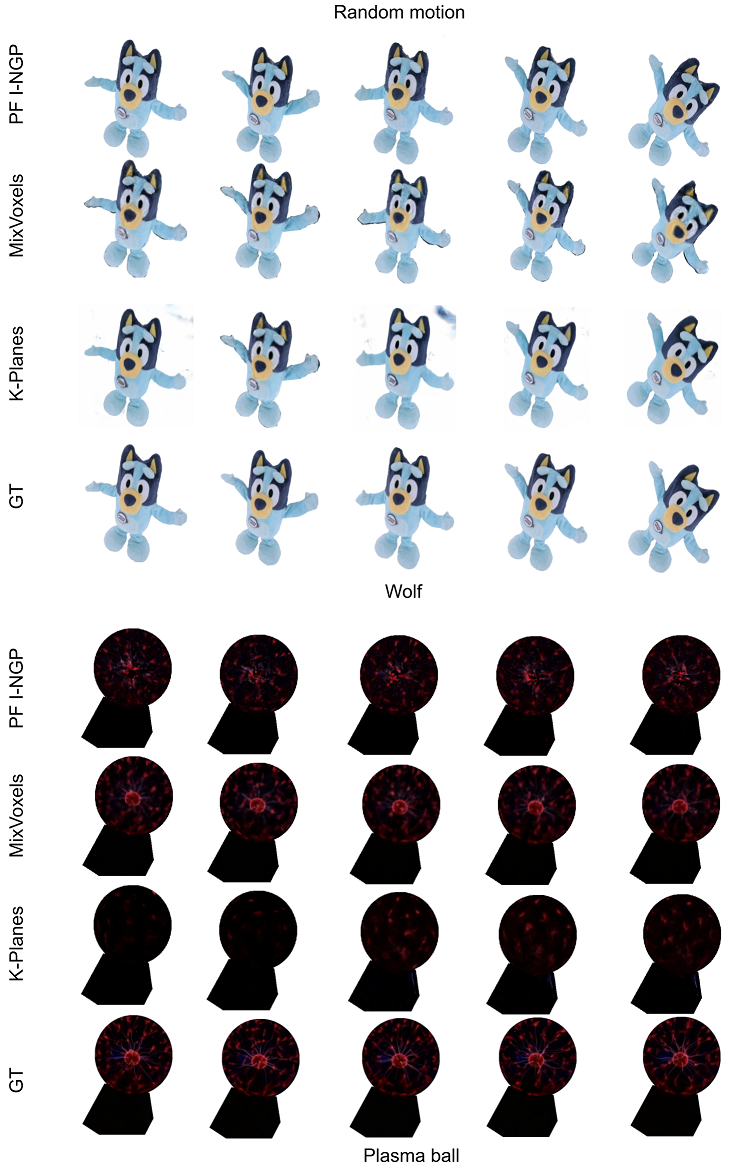}
  \end{subfigure}
  
  \caption{Test view reconstruction and ground truth of two random motion objects: toy wolf and plasma ball. Top: Both MixVoxels and K-Planes capture the toy's motion, but MixVoxels generates some artifacts and sometimes fails to capture parts of the ear and foot of the wolf, and K-Planes contains a few floaters in some frames. Bottom: MixVoxels captures the currents in the plasma ball surprisingly well, while K-Planes and PF I-NGP completely fail. Notably, the plasma ball is captured from a darker environment, so we use the ground truth mask to turn the background into white color for visually pleasing purposes.}
  \label{fig:random}
\end{figure*}

All baselines perform slightly better on detailed motions. 
However, a gap exists between MixVoxels and PF I-NGP's quantitative results, suggesting that MixVoxels can capture the static background but struggles to capture detailed motion (e.g. the second hand of the Clock in \Cref{fig:detailed} disappears). 
Unlike MixVoxels, K-Planes partially reconstructs the second hand but fails to reconstruct the static part well (e.g., the clock's body).
This leads to a larger performance gap between K-Planes and PF I-NGP.
Slow and repetitive motions are the two categories that MixVoxels' results are the closest to PF I-NGP's while the performance gaps are larger in fast and random motions. 
In other words, MixVoxels can successfully capture dynamic information when the motion is continuous and gradual but cannot generalize well to drastic motion. 
For example, the Horse in \Cref{fig:slow} and the Penguin in \Cref{fig:repetitive} are clean, while the Blue Car in \Cref{fig:fast} and the Wolf in \Cref{fig:random} miss fine-grained details or parts.
Fast and random motions are the two categories that K-Planes' results are the closest to PF I-NGP's while the performance gaps are larger in slow and repetitive motions. 
In other words, K-Planes can construct scenes of drastic motion with fewer artifacts but may produce more artifacts, such as floaters for mild motion.
For instance, the Dog in \Cref{fig:fast} and the Wolf in \Cref{fig:random} contain fewer floaters and artifacts than the Horse in \Cref{fig:slow} and the Stirling Engine in \Cref{fig:repetitive}.
Together with the quantitative results (\Cref{tab:dynamic_benchmark_all_objects}), the visualization results suggest that PF I-NGP can successfully capture most of the scene, whereas Mixvoxels and K-Planes struggle to generalize to different types of motions and may need hyperparameter tuning to fit scenes with different motion types.

\begin{table*}[tbh!]
  \centering
  \begin{tabular}{cccccc|c}
    \toprule
    Scene & Slow & Fast & Detailed & Repetitive & Random & Size L$\times$W$\times$H (cm) \\
    \midrule
    blue car & \xmark & \checkmark & \xmark & \checkmark & \xmark & 23.20\ $\times$\ \ \ 9.50\ $\times$\ \ \ 5.20 \\
    bunny & \xmark & \checkmark & \xmark & \checkmark & \xmark & 33.02\ $\times$\ 27.94\ $\times$\ 19.05\\
    clock & \checkmark & \xmark & \checkmark & \xmark & \xmark & 30.48\ $\times$\ 30.48$\ \times\ \ \ $4.57\\
    dog & \xmark & \checkmark & \xmark & \xmark & \checkmark & 27.00\ $\times$\ 11.99\ $\times$\ 27.00\\
    horse & \checkmark & \xmark & \xmark & \checkmark & \xmark & 11.50\ $\times$\ \ \ 7.60\ $\times$\ 13.50\\
    hourglass & \checkmark & \xmark & \checkmark & \xmark & \checkmark & \ \ 7.00\ $\times$\ \ \ 7.00\ $\times$\ 17.00\\
    k1 double punch & \xmark & \checkmark & \xmark & \checkmark & \xmark & \ \ 9.80\ $\times$\ 17.20\ $\times$\ 35.00\\
    k1 handstand & \xmark & \checkmark & \xmark & \xmark & \xmark & \ \ 9.80\ $\times$\ 17.20\ $\times$\ \ 35.00\\
    k1 push up & \xmark & \checkmark & \xmark & \checkmark & \xmark & \ \ 9.80\ $\times$\ 17.20\ $\times$\ 35.00\\
    music box & \checkmark & \xmark & \xmark & \checkmark & \xmark & 10.80\ $\times$\ 10.80\ $\times$\ 12.60\\
    penguin & \xmark & \xmark & \xmark & \checkmark & \xmark & 24.69\ $\times$\ 27.00\ $\times$\ 13.77\\
    plasma ball & \xmark & \checkmark & \checkmark & \xmark & \checkmark & 15.00\ $\times$\ 15.00\ $\times$\ 24.10\\
    plasma ball clip & \xmark & \checkmark & \checkmark & \xmark & \checkmark & 15.00\ $\times$\ 15.00\ $\times$\ 24.10\\
    red car & \xmark & \checkmark & \xmark & \checkmark & \xmark & 22.50\ $\times$\ \ \ 9.20\ $\times$\ \ \ 5.20\\
    stirling & \xmark & \checkmark & \checkmark & \checkmark & \xmark & \ \ 9.00\ $\times$\ \ \ 9.00\ $\times$\ 13.00\\
    tornado & \xmark & \checkmark & \xmark & \checkmark & \xmark & 11.00\ $\times$\ 11.00\ $\times$\ 27.50\\
    trex & \xmark & \checkmark & \xmark & \xmark & \xmark & 30.50\ $\times$\ \ \ 7.50\ $\times$\ 20.50\\
    truck & \xmark & \checkmark & \xmark & \xmark & \xmark & 17.80\ $\times$\ \ \ 7.10\ $\times$\ \ \ 7.30\\
    wall-e & \xmark & \checkmark & \xmark & \xmark & \xmark & 35.56\ $\times$\ 20.32\ $\times$\ 27.94\\
    wolf & \xmark & \xmark & \xmark & \checkmark & \checkmark & 17.78\ $\times$\ 12.70\ $\times$\ 35.56\\
    world globe & \checkmark & \xmark & \xmark & \checkmark & \xmark & 13.97\ $\times$\ 13.97\ $\times$\ 19.81\\   
    \bottomrule\\
  \end{tabular}
  \caption{Motion types and object size (cm) of each dynamic object. Objects are split into 5 overlapping categories: slow, fast, detailed, repetitive, and random motion. }
    \label{tab:dynamic_object_types}
\end{table*}

\begin{figure*}[t!]
  \centering
  \begin{subfigure}[b]{\textwidth}
    \includegraphics[width=\textwidth]{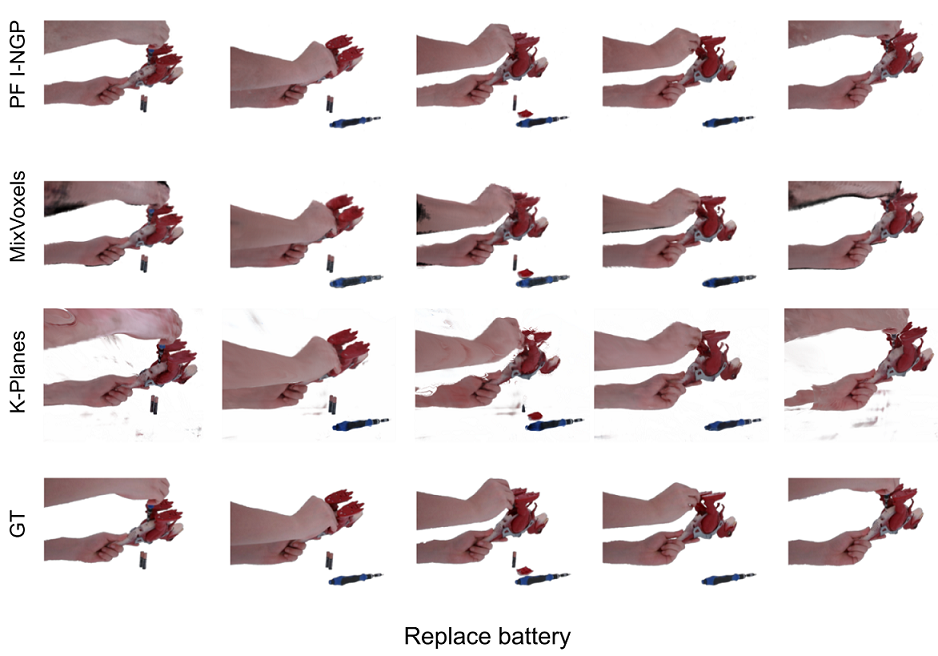}
  \end{subfigure}
  
  \caption{Interaction scene showing the motion of replacing a toy Trex's battery. The sequence contains a series of realistic motions. Both PF I-NGP and MixVoxels correctly reconstruct the scene. K-Planes distorts the hand in the first column and generates many floaters.}
  \label{fig:battery}
\end{figure*}

\begin{figure*}[t!]
  \centering
  \begin{subfigure}[b]{\textwidth}
    \includegraphics[width=\textwidth]{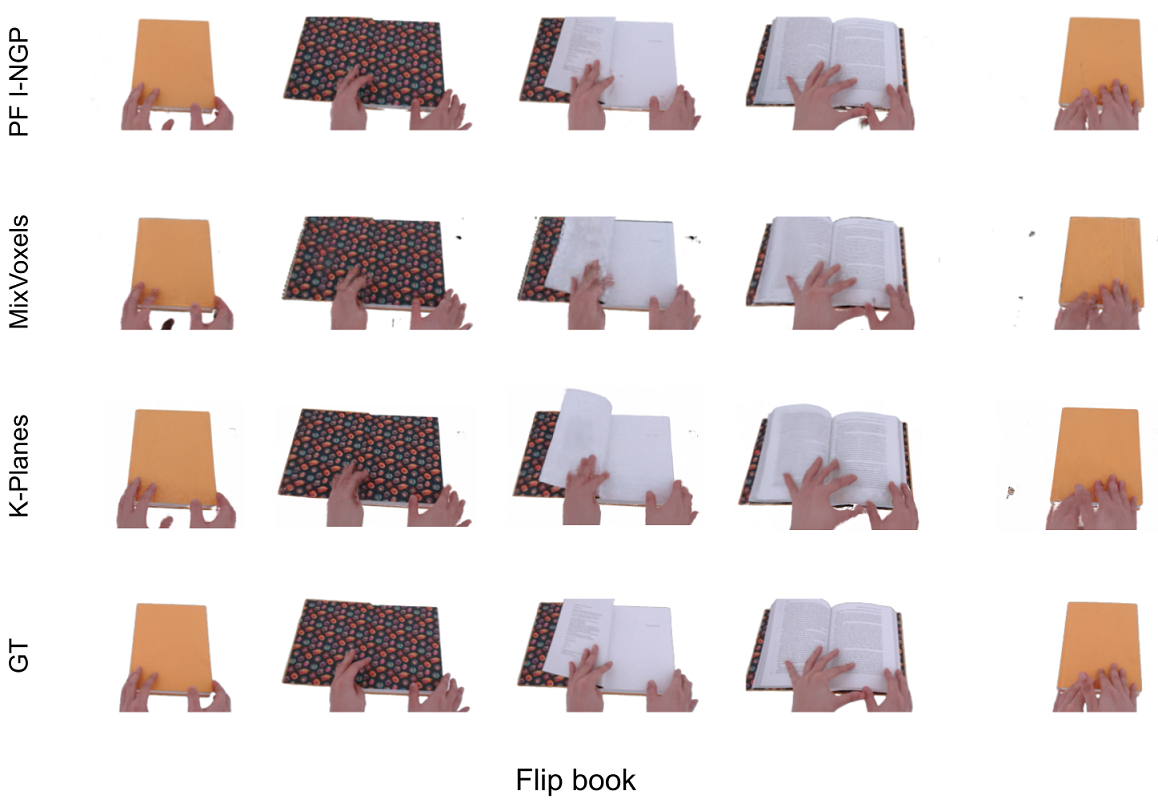}
  \end{subfigure}
  
  \caption{Interaction scene showing the motion of flipping through a book. PF I-NGP generates some artifacts around the bottom of the book in the second and fourth columns. MixVoxels generates blurry book pages in the third column. K-Planes totally misses the texts in the third column.}
  \label{fig:flip book}
\end{figure*}

\begin{figure*}[t!]
  \centering
  \begin{subfigure}[b]{\textwidth}
    \includegraphics[width=\textwidth]{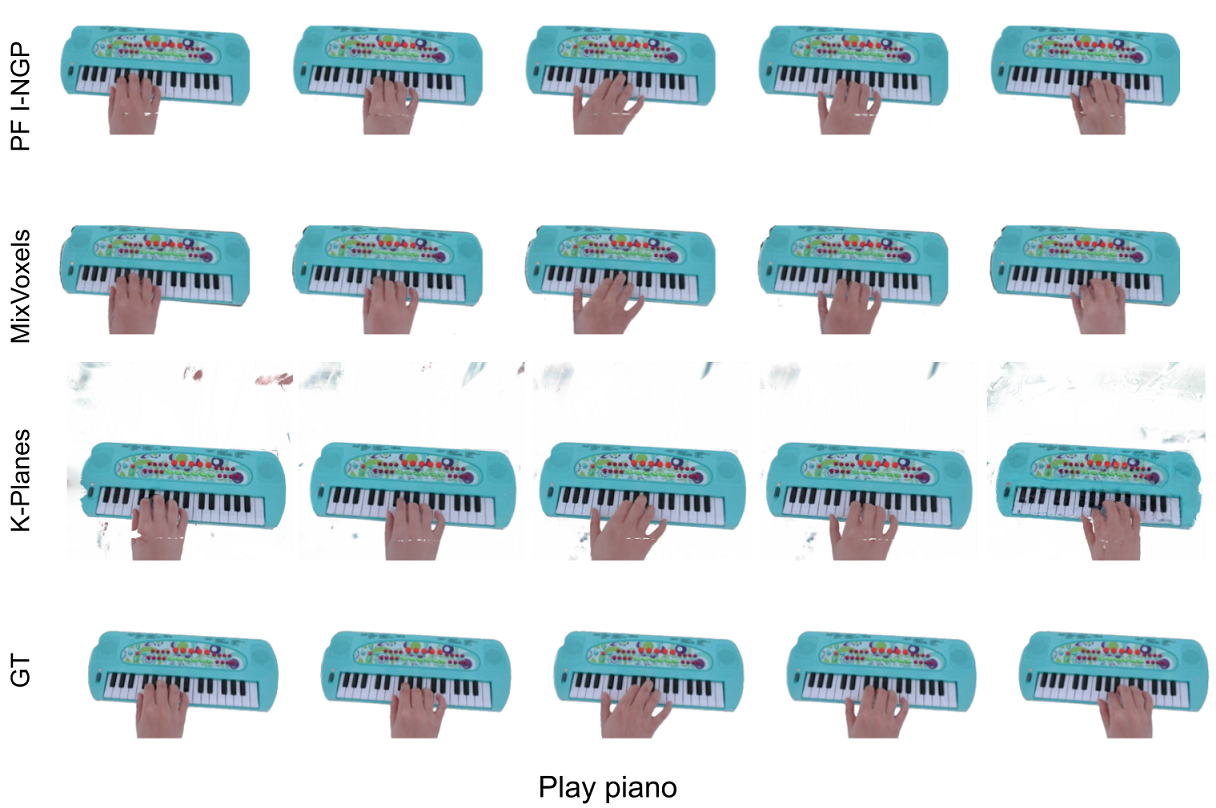}
  \end{subfigure}
  
  \caption{Interaction scene showing a hand playing a toy piano. Both PF I-NGP and MixVoxels capture the motion successfully, but PF I-NGP produces some white scratches on the back of the hand. K-Planes produces white scratches on the back of the hand and floaters in the background.}
  \label{fig:piano}
\end{figure*}

\begin{figure*}[t!]
  \centering
  \begin{subfigure}[b]{\textwidth}
    \includegraphics[width=\textwidth]{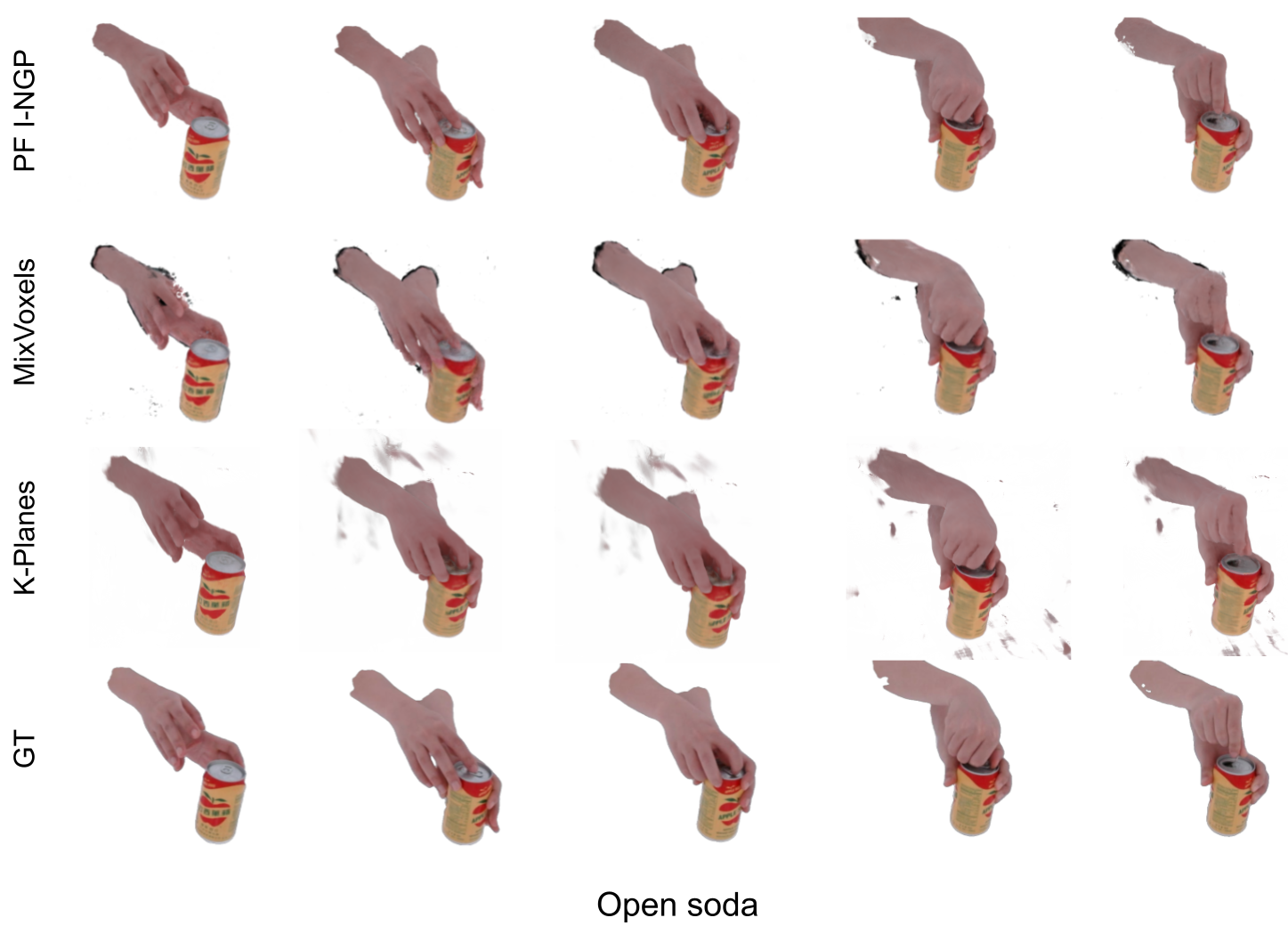}
  \end{subfigure}
  
  \caption{Interaction scene showing the process of opening a can of soda. MixVoxels generates a few floaters around the hand in the first column. K-Planes generates many floaters in the background.}
  \label{fig:soda}
\end{figure*}

\begin{figure*}[t!]
  \centering
  \begin{subfigure}[b]{\textwidth}
    \includegraphics[width=\textwidth]{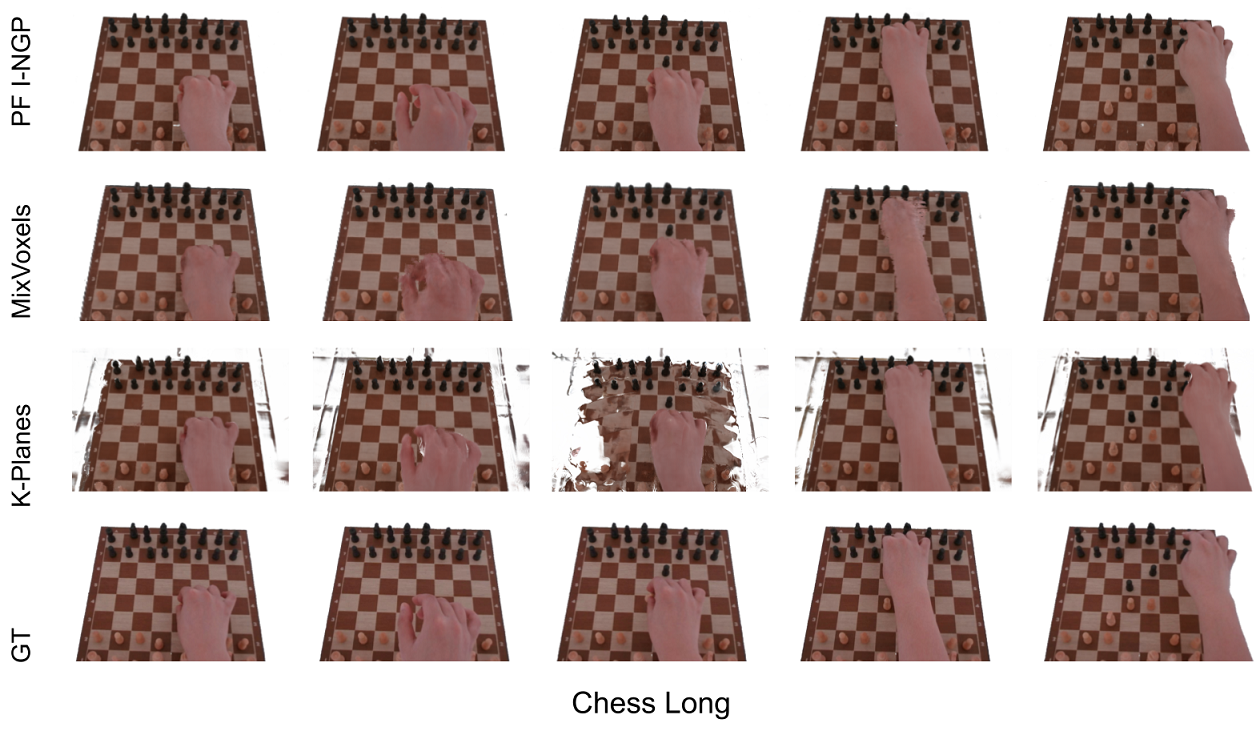}
  \end{subfigure}
  
  \caption{Long-duration scene showing people playing chess. Both PF I-NGP and MixVoxels can reconstruct the chessboard and chess well, while K-Planes struggles at these static parts. For dynamic parts, both PF I-NGP and K-Planes can capture the hand clearly, while MixVoxels cannot.}
  \label{fig:long_chess}
\end{figure*}

\parahead{Interaction Results}
Our interaction scenes, which cover several human daily activities such as flipping a book in \Cref{fig:flip book}, require the model to capture a sequence of realistic motions.
For example, the Battery scene in \Cref{fig:battery} contains the motion of using a screwdriver to open the toy's battery cover, putting in the batteries, assembling the cover back, and turning on the toy.
We consider the interaction data as more challenging cases for neural radiance fields because of the occlusion and complex motion from the hands.
\Cref{tab:dynamic_benchmark_interactions} demonstrates all results of PF I-NGP, MixVoxels, and K-Planes on interaction scenes. 
Overall, the performances of these three baselines are similar across interaction scenes.
PF I-NGP performs robustly across most of the scenes while failing to totally reconstruct the cover of the book in \Cref{fig:flip book} and produces scratches on the hand in \Cref{fig:piano}.
MixVoxels reconstructs blurrier texts in \Cref{fig:flip book}, and a few floaters around the hand in \Cref{fig:soda}.
K-Planes reconstructs many floaters around the hand and objects in ~\Cref{fig:battery,fig:piano,fig:soda}, misses the text in ~\Cref{fig:flip book}, distorts the hand in \Cref{fig:battery}, and produces white scratches on the hand in \Cref{fig:piano}.
We hope that our interaction dynamic dataset can open a new direction and provide a new understanding for hand object interaction tasks~\cite{blukis2022neural} in the future.
Notably, visualization results show black artifacts in \Cref{fig:battery} and \Cref{fig:soda} in MixVoxels renderings. 
This effect is not observable in numerical results because all models except K-Planes are trained and evaluated with black backgrounds. 
We used white backgrounds for rendering to better visualize the results.

\begin{table*}[tbh!]
  \centering
\scalebox{0.73}{
  \begin{tabular}{cccccc}
    \toprule
    Scene & \shortstack{PF I-NGP / MixVoxels / K-Planes\\PSNR$\uparrow$} & \shortstack{PF I-NGP / MixVoxels / K-Planes\\SSIM$\uparrow$} & \shortstack{PF I-NGP / MixVoxels / K-Planes\\LPIPS$\downarrow$} & \shortstack{PF I-NGP / MixVoxels / K-Planes\\JOD$\uparrow$} \\
    \midrule
    blue car & 29.833 / 29.485 / 28.304 & 0.957 / 0.955 / 0.962 & 0.047 / 0.049 / 0.059 & 7.951 / 8.097 / 8.581 \\
    bunny & 26.491 / 24.983 / 27.244 & 0.941 / 0.928 / 0.935 & 0.085 / 0.098 / 0.176 & 7.905 / 7.407 / 7.790 \\
    clock & 28.943 / 28.682 / 21.972 & 0.935 / 0.934 / 0.868 & 0.108 / 0.100 / 0.299 & 7.810 / 9.026 / 6.833 \\
    dog & 25.463 / 23.233 / 29.483 & 0.949 / 0.929 / 0.949 & 0.085 / 0.100 / 0.143 & 7.754 / 6.350 / 8.405 \\
    horse & 31.869 / 31.724 / 25.940 & 0.982 / 0.980 / 0.878 & 0.023 / 0.023 / 0.333 & 8.736 / 8.910 / 5.816 \\
    hourglass & 27.244 / 27.468 / 27.559 & 0.970 / 0.976 / 0.930 & 0.054 / 0.027 / 0.261 & 2.572 / 3.100 / 7.477 \\
    k1 double punch & 27.422 / 27.208 / 23.190 & 0.938 / 0.938 / 0.917 & 0.070 / 0.068 / 0.202 & 6.733 / 6.421 / 6.832 \\
    k1 handstand & 27.631 / 26.795 / 23.178 & 0.936 / 0.930  / 0.906 & 0.072 / 0.078 / 0.180 & 7.233 / 7.377 / 5.945 \\
    k1 push up & 27.393 / 27.326 / 22.345 & 0.936 / 0.935 / 0.889 & 0.072 / 0.072 / 0.230 & 6.886 / 7.439 / 5.006 \\
    music box & 32.225 / 32.129 / 24.862 & 0.980 / 0.979 / 0.871 & 0.031 / 0.028 / 0.329 & 8.444 / 8.636 / 5.004 \\
    penguin & 27.034 / 26.675 / 28.504 & 0.950 / 0.950 / 0.949 & 0.074 / 0.068 / 0.165 & 8.182 / 8.291 / 8.338 \\
    plasma ball & 33.422 / 36.102 / 29.145 & 0.944 / 0.955 / 0.857 & 0.072 / 0.060 / 0.273 & 7.368 / 6.372 / 5.048 \\
    plasma ball clip & 46.476 /~~~~\NA~~~~~/ 41.437 & 0.968 /~~~\NA~~~~/ 0.940 & 0.049 /~~~\NA~~~/ 0.108 & 8.139 /~~~\NA~~~/ 5.502\\
    red car & 30.844 / 30.342 / 28.626 & 0.961 / 0.960 / 0.962 & 0.046 / 0.050 / 0.111 & 8.020 / 8.161 / 8.352 \\
    stirling & 29.473 / 28.744 / 19.415 & 0.966 / 0.963 / 0.810 & 0.045 / 0.037 / 0.372 & 7.808 / 7.868 / 3.550 \\
    tornado & 28.629 / 28.825 / 24.427 & 0.965 / 0.966 / 0.886 & 0.045 / 0.043 / 0.309 & 6.398 / 6.815 / 6.164 \\
    trex & 28.496 / 28.057 / 24.976 & 0.948 / 0.947 / 0.935 & 0.056 / 0.058 / 0.148 & 8.257 / 8.172 / 6.266 \\
    truck & 31.033 / 30.654 / 30.431 & 0.969 / 0.967 / 0.973 & 0.038 / 0.044 / 0.064 & 8.418 / 8.412 / 8.849 \\
    wall-e & 28.164 / 27.263 / 25.694 & 0.934 / 0.923 / 0.936 & 0.085 / 0.114 / 0.134 & 7.597 / 7.399 / 7.933 \\
    wolf & 25.341 / 24.764 / 26.683 & 0.940 / 0.935 / 0.932 & 0.089 / 0.094 / 0.183 & 7.855 / 7.810 / 6.800 \\
    world globe & 28.529 / 28.175 / 25.560 & 0.953 / 0.955 / 0.874 & 0.052 / 0.047 / 0.322 & 8.063 / 8.243 / 5.372 \\   
    \bottomrule\\
  \end{tabular}
  }
  \caption{Rendering quality of all dynamic objects using PF I-NGP~\cite{mueller2022instant}, MixVoxels~\cite{wang2023mixed} and K-Planes~\cite{kplanes_2023}. PF I-NGP and MixVoxels are evaluated with black backgrounds. K-Planes is evaluated with white backgrounds.} 
    \label{tab:dynamic_benchmark_all_objects}
\end{table*}

\begin{figure*}[t!]
  \centering
  \begin{subfigure}[b]{\textwidth}
    \includegraphics[width=\textwidth]{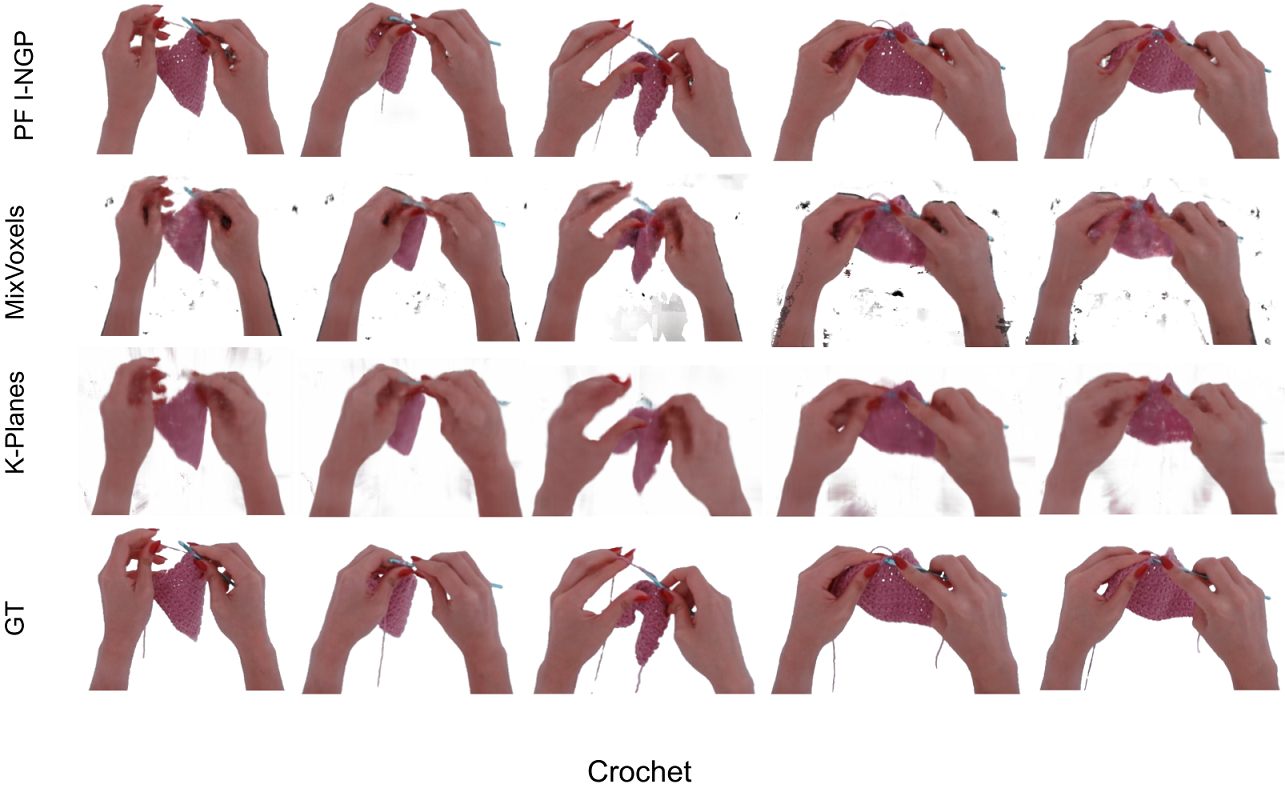}
  \end{subfigure}
  
  \caption{Long-duration scene showing a person crocheting. PF I-NGP can capture the holes in the fabrics, while MixVoxels and K-Planes smooth the fabrics.}
  \label{fig:crochet}
\end{figure*}

\parahead{Long duration Results}
Although our object and interaction datasets contain several long videos, we propose a long-duration dataset that only includes videos that are at least 2 minutes long (see \Cref{fig:dist}). 
\Cref{tab:dynamic_benchmark_long} details the rendering quality of PF I-NGP, MixVoxels, and K-Planes on each scene of the long-duration data.
Surprisingly, MixVoxls outperforms PF I-NGP in many scenes of the dataset.
This is because many scenes of the dataset contain large static parts such as a chessboard in the Chess Long, and the tray in the Jenga Long, Legos, Origami, Painting, and Puzzle.
MixVoxels is good at maintaining the consistency of the static parts across frames, while PF I-NGP cannot (see \Cref{fig:tmp_inc}).
\Cref{fig:crochet} demonstrates that PF I-NGP still provides more fine-grained details, such as the holes of fabrics in Crochet, than MixVoxels and K-Planes. 
\Cref{fig:long_chess} shows that MixVoxels performs better at static parts like the chessboard while K-Planes performs better at dynamic parts such as the hand.

\section{Dynamic Dataset Experiments}

In this section, we discuss the visualization results for each experiment in Section 5.2 of the paper. 

\begin{figure*}[t!]
  \centering
  \begin{subfigure}[b]{\textwidth}
    \includegraphics[width=\textwidth]{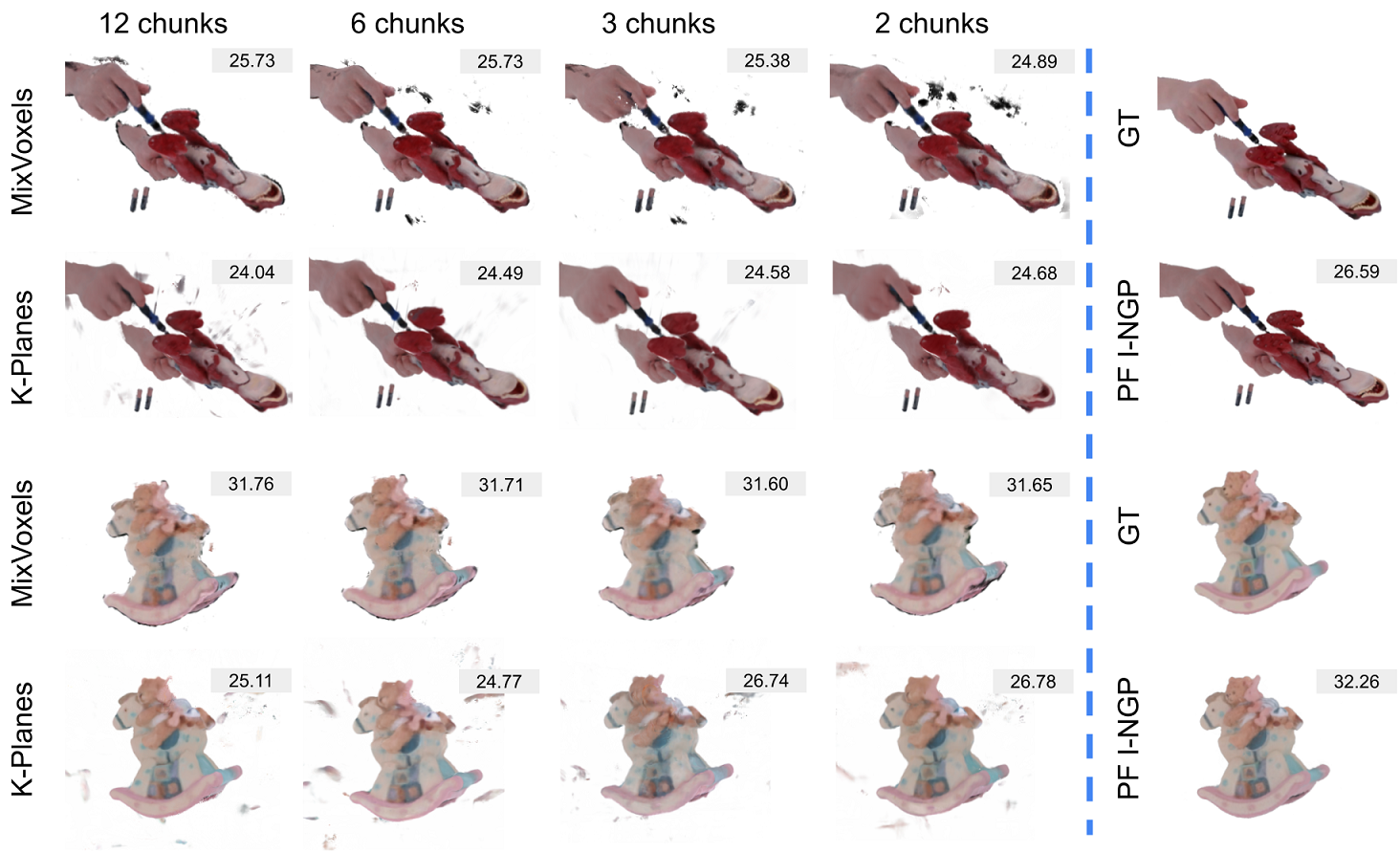}
  \end{subfigure}
  
  \caption{Visualization results of three baselines trained with sequences that split into 12, 6, 3, and 2 chunks. More chunks indicate a shorter temporal length per model and less temporal information. MixVoxels is less noise in Replace Battery, and the bear's eyes in Horse are more apparent when the amount of chunks increases. Unlike MixVoxels, K-Planes generates fewer floaters around the object when the amount of chunks decreases.}
  \label{fig:exp_b}
\end{figure*}

\parahead{Temporal Information}
\Cref{fig:exp_b} demonstrates the visualization results of MixVoxels and K-Planes when trained with different numbers of chunks. 
Notably, more chunks indicate a shorter temporal length per model and less temporal information.
Visualization results of MixVoxels show less noise when the number of chunks increases, while K-Planes' results show the opposite.
Hence, the visualization results support that MixVoxels prefers a shorter temporal length per model, while K-Planes prefers a longer temporal length per model.

\begin{figure*}[t!]
  \centering
  \begin{subfigure}[b]{\textwidth}
    \includegraphics[width=\textwidth]{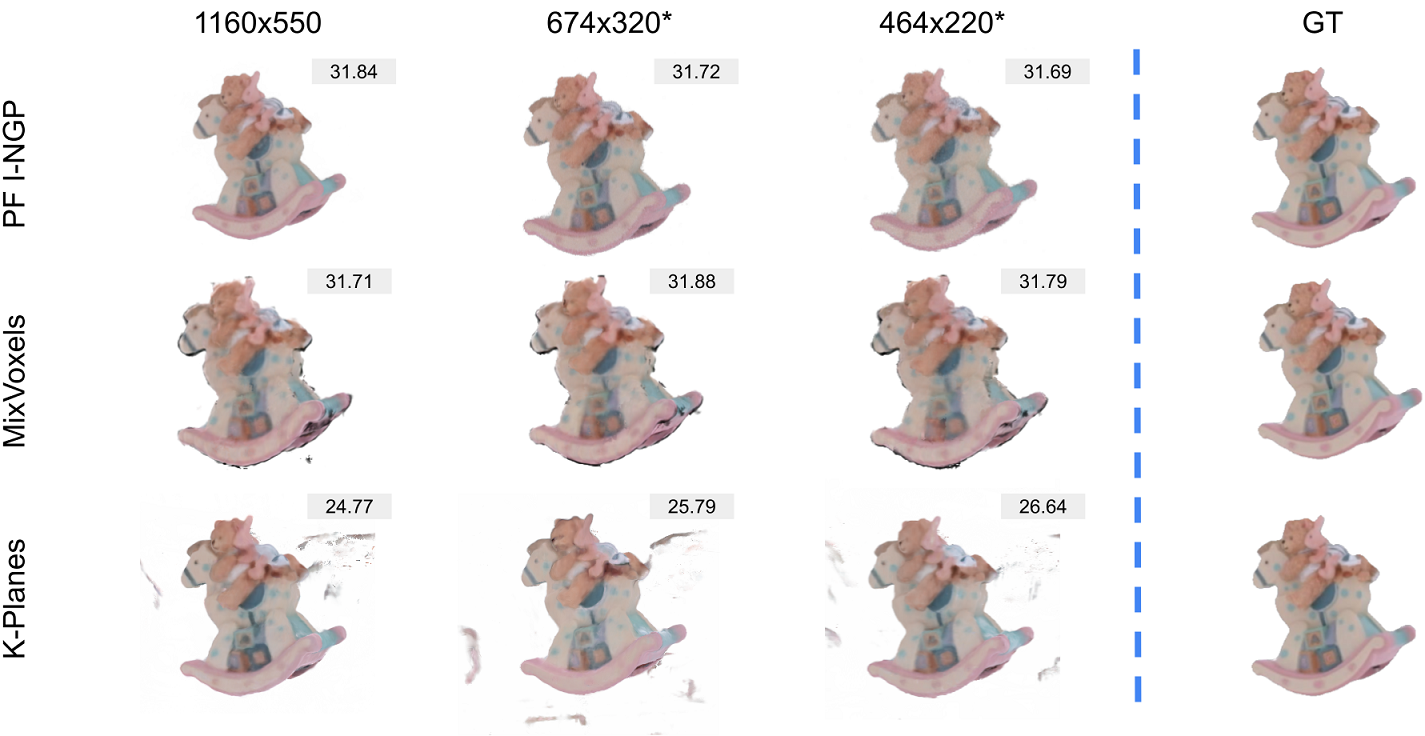}
  \end{subfigure}
  
  \caption{Visualization results of PF I-NGP, MixVoxels, and K-Planes trained with images in different resolutions. * indicates that we spatially interpolate the rendering results to 1160$\times$550 during testing. The visualization results of PF I-NGP and MixVoxels are similar across three settings in Horse, but the stripes of the bunny's clothes are not well reconstructed in 464$\times$200 (better zoom in to see the details). The visualization results of K-Planes contain fewer floaters when the resolution decreases.}
  \label{fig:exp_h}
\end{figure*}

\parahead{Spatial Information}
From \Cref{fig:exp_h}, firstly, we notice that the spatial interpolation ability of the neural radiance field is impressive. 
The visualization of PF I-NGP and MixVoxels are almost the same across three resolutions, but they start missing fine-grained details, such as the stripes of the Bunny's clothes in the lowest resolution setting. 
This indicates that the performance drop won't happen if the resolution is acceptable for neural radiance fields.
Secondly, we found that the reconstruction results of MixVoxels and K-Planes do not miss any parts of the object but are similarly blurry across all three settings.
This suggests that current dynamic NeRFs may be biased towards capturing the shape of moving objects, potentially sacrificing the ability to capture details.
Finally, the floaters generated from K-Planes start disappearing when the resolution gets lower.
This is because K-Planes can revisit the same training samples frequently under the same training setting.

\begin{table*}[tbh!]
  \centering
  \scalebox{0.75}{
  \begin{tabular}{cccccc}
    \toprule
    Scene & PF I-NGP / Mixvoxels / K-Planes & PF I-NGP / Mixvoxels /K-Planes & PF I-NGP / Mixvoxels / K-Planes & PF I-NGP / Mixvoxels / K-Planes \\& PSNR$\uparrow$ & SSIM$\uparrow$ & LPIPS$\downarrow$ & JOD$\uparrow$\\
    \midrule
     battery & 26.828 / 25.805 / 24.351 & 0.931 / 0.902 / 0.923 & 0.088 / 0.128 / 0.169 & 7.483 / 6.729 / 7.404 \\
     chess & 22.945 / 20.799 / 22.274 & 0.821 / 0.807 / 0.865 & 0.215 / 0.233 / 0.267 & 6.756 / 5.932 / 7.691 \\
     drum & 24.662 / 23.784 / 22.299 & 0.903 / 0.894 / 0.880 & 0.136 / 0.148 / 0.253 &  7.703 / 6.663 / 7.199\\
     flip book & 26.303 / 24.367 / 25.826 & 0.928 / 0.904 / 0.920 & 0.120 / 0.150 / 0.198 & 7.347 / 5.750 / 8.093 \\
     jenga & 29.683 / 28.884 / 30.992 & 0.972 / 0.962 / 0.973 & 0.046 / 0.063 / 0.087 & 8.583 / 8.281 / 8.284 \\
     keyboard mouse & 29.635 / 29.153 / 24.734 & 0.926 / 0.926 / 0.915 & 0.102 / 0.101 / 0.204 & 7.808 / 7.777 / 6.910 \\
     kindle & 29.556 / 28.585 / 25.796 & 0.958 / 0.951 / 0.943 & 0.069 / 0.087 / 0.169 & 8.237 / 7.969 / 6.816 \\
     maracas & 26.083 / 26.300 / 27.352 & 0.953 / 0.949 / 0.960 & 0.072 / 0.081 / 0.085 & 7.659 / 7.575 / 8.312 \\
     pan & 27.392 / 26.759 / 24.803 & 0.935 / 0.918 / 0.937 & 0.094 / 0.116 / 0.153 & 7.003 / 6.996 / 7.532 \\
     peel apple & 27.270 / 27.205 / 26.108 & 0.939 / 0.942 / 0.934 & 0.086 / 0.089 / 0.184 & 7.843 / 7.728 / 7.595 \\
     piano & 26.824 / 26.097 / 23.879 & 0.929 / 0.925 / 0.886 & 0.104 / 0.099 / 0.257 & 7.719 / 8.438 / 7.013 \\
     poker & 27.786 / 27.736 / 29.137 & 0.958 / 0.954 / 0.960 & 0.062 / 0.068 / 0.095 & 8.298 / 8.143 / 8.409 \\
     pour salt & 25.845 / 25.729 / 24.702 & 0.919 / 0.919 / 0.903 & 0.104 / 0.111 / 0.225 & 7.714 / 7.311 / 7.622 \\
     pour tea & 26.071 / 25.775 / 27.626 & 0.946 / 0.941 / 0.950 & 0.089 / 0.094 / 0.144 & 7.447 / 6.951 / 7.607 \\
     put candy & 28.189 / 27.360 / 25.870 & 0.950 / 0.943 / 0.940 & 0.071 / 0.078 / 0.162 & 7.906 / 7.836 / 6.737 \\
     put fruit & 27.129 / 26.614 / 26.546 & 0.935 / 0.931 / 0.938 & 0.089 / 0.097 / 0.141 & 7.815 / 7.389 / 7.456 \\
     scissor & 25.346 / 25.090 / 25.883 & 0.944 / 0.937 / 0.936 & 0.076 / 0.086 / 0.168 & 7.854 / 7.563 / 7.685 \\
     slice apple & 26.026 / 25.014 / 26.692 & 0.951 / 0.940 / 0.939 & 0.106 / 0.118 / 0.218 & 7.829 / 7.515 / 7.948 \\
     soda & 28.780 / 28.727 / 27.115 & 0.964 / 0.959 / 0.936 & 0.059 / 0.067 / 0.184 & 8.322 / 8.091 / 7.704 \\
     tambourine & 27.985 / 27.624 / 27.482 & 0.972 / 0.965 / 0.968 & 0.044 / 0.049 / 0.100 & 7.634 / 7.455 / 7.227 \\
     tea & 27.410 / 26.808 / 28.202 & 0.956 / 0.946 / 0.962 & 0.073 / 0.088 / 0.099 & 7.723 / 7.268 / 7.946 \\
     unlock & 28.649 / 29.275 / 29.967 & 0.971 / 0.966 / 0.974 & 0.045 / 0.058 / 0.070 & 8.158 / 8.243 / 8.441 \\
     writing 1 & 24.086 / 25.145 / 25.630 & 0.916 / 0.926 / 0.926 & 0.158 / 0.163 / 0.236 & 6.725 / 7.789 / 7.222 \\
     writing 2 & 24.345 / 25.613 / 25.782 & 0.930 / 0.934 / 0.926 & 0.146 / 0.148 / 0.242 & 6.490 / 7.983 / 7.604 \\
     xylophone & 27.334 / 26.951 / 25.449 & 0.950 / 0.948 / 0.916 & 0.068 / 0.066 / 0.206 & 7.950 / 8.103 / 7.094 \\
    \bottomrule\\
  \end{tabular}
  }
  \caption{Rendering quality of all dynamic interaction scenes using PF I-NGP\cite{mueller2022instant}, MixVoxels\cite{wang2023mixed} and K-Planes\cite{kplanes_2023}. PF I-NGP and MixVoxels are evaluated with black backgrounds. K-Planes is evaluated with white backgrounds.}
    \label{tab:dynamic_benchmark_interactions}
\end{table*}

\begin{figure*}[t!]
  \centering
  \begin{subfigure}[b]{\textwidth}
    \includegraphics[width=\textwidth]{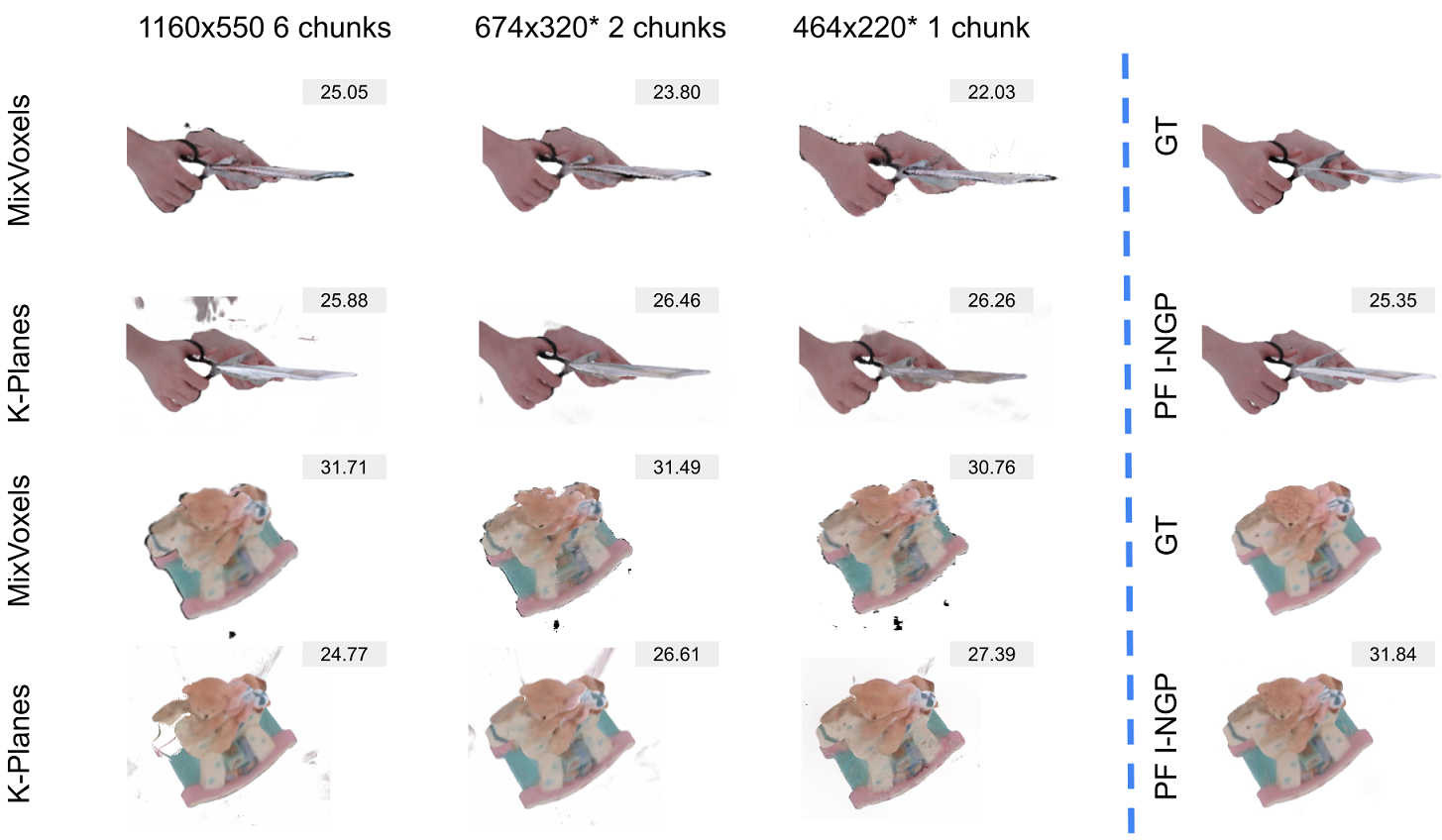}
  \end{subfigure}
  
  \caption{Controlling the spatial resolution and amount of chunks simultaneously does not break the property of MixVoxels and K-Planes.}
  \label{fig:exp_c}
\end{figure*}

\parahead{Spatial and Temporal Information}
In \Cref{fig:exp_c}, the reconstruction results of MixVoxels are blurrier (the hand in Scissor) and incomplete (the bear's ears in Horse) on the lowest resolution and more temporal information settings.
Although the reconstruction results of K-Planes are slightly blurrier (the hand in Scissor), the reconstructions are complete (the Horse's head) and have fewer floaters with the lowest resolution and more temporal information setting.
These match our previous experiment results where we only control one of these factors at a time.

\section{Dataset Justification}
In this section, we discuss the visualization results of \brics (\textit{All-view}), \brics with just two panels (\textit{Forward}), and \brics with fewer cameras (\textit{Multi-view}) to support that multi-view~360$^\circ$ setting with enough cameras can lead to the most completed reconstruction.

\begin{figure*}[t!]
  \centering
  \begin{subfigure}[b]{\textwidth}
    \includegraphics[width=\textwidth]{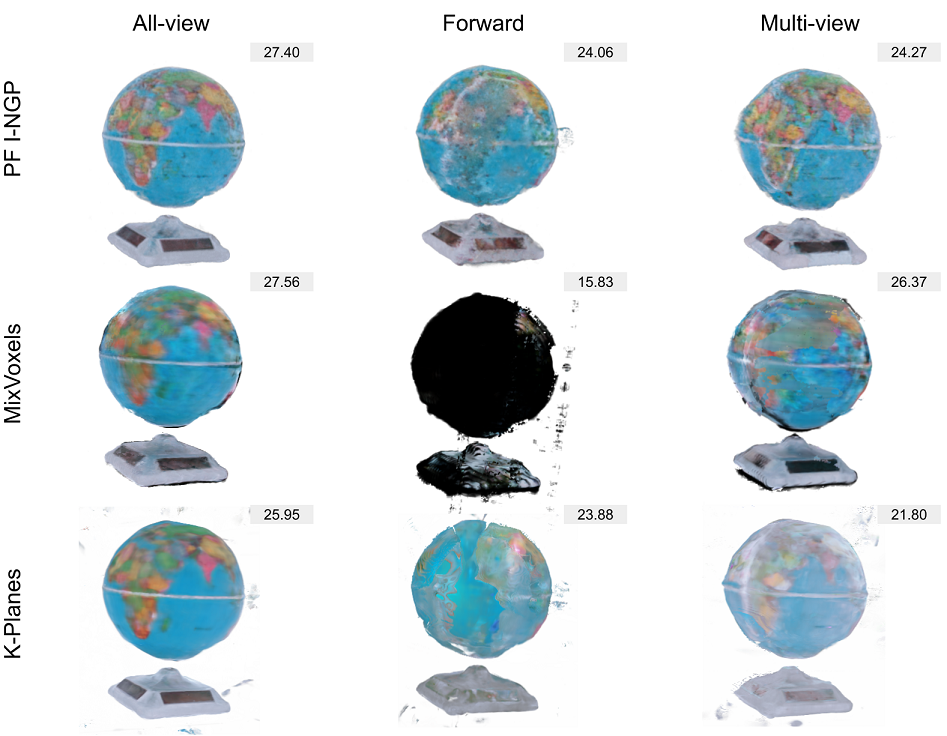}
  \end{subfigure}
  
  \caption{The visualization results of PF I-NGP, MixVoxels, and K-Planes trained with World Globe captured from \brics (\textit{All-view}), two panels of \brics (\textit{Forward}), and \brics with fewer cameras (\textit{Multi-view}). We render the occluded view of \textit{Forward} to demonstrate that the Multi-view~360$^\circ$ setting with enough cameras can provide the most comprehensive reconstruction. Both PF I-NGP and K-Planes can render the occluded view with roughly similar RGB colors, while MixVoxels renders the occluded view with black background color.}
  \label{fig:exp_g_1}
\end{figure*}

\begin{figure*}[t!]
  \centering
  \begin{subfigure}[b]{\textwidth}
    \includegraphics[width=\textwidth]{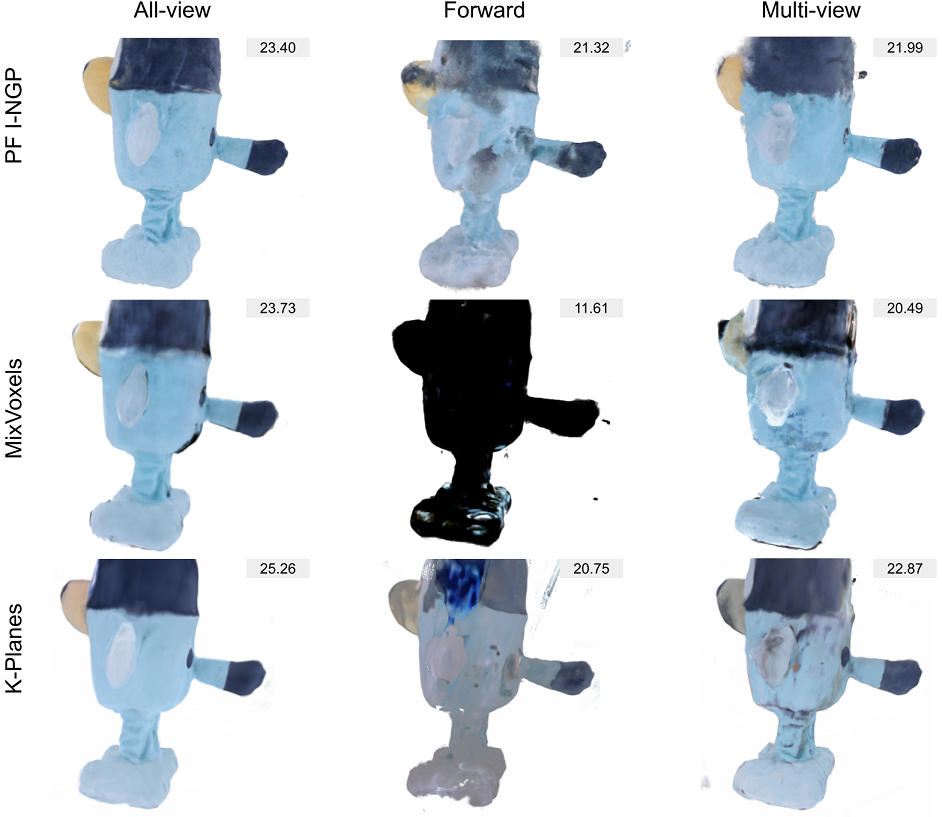}
  \end{subfigure}
  
  \caption{The visualization results of PF I-NGP, MixVoxels, and K-Planes trained with Wolf captured from \brics (\textit{All-view}), two panels of \brics (\textit{Forward}), and \brics with fewer cameras (\textit{Multi-view}). We render the occluded view of \textit{Forward} to demonstrate that the Multi-view~360$^\circ$ setting with enough cameras can provide the most comprehensive reconstruction. Both PF I-NGP and K-Planes can render the occluded view with roughly similar RGB colors, while MixVoxels renders the occluded view with black background color.}
  \label{fig:exp_g_2}
\end{figure*}

\begin{figure*}[t!]
  \centering
  \begin{subfigure}[b]{\textwidth}
    \includegraphics[width=\textwidth]{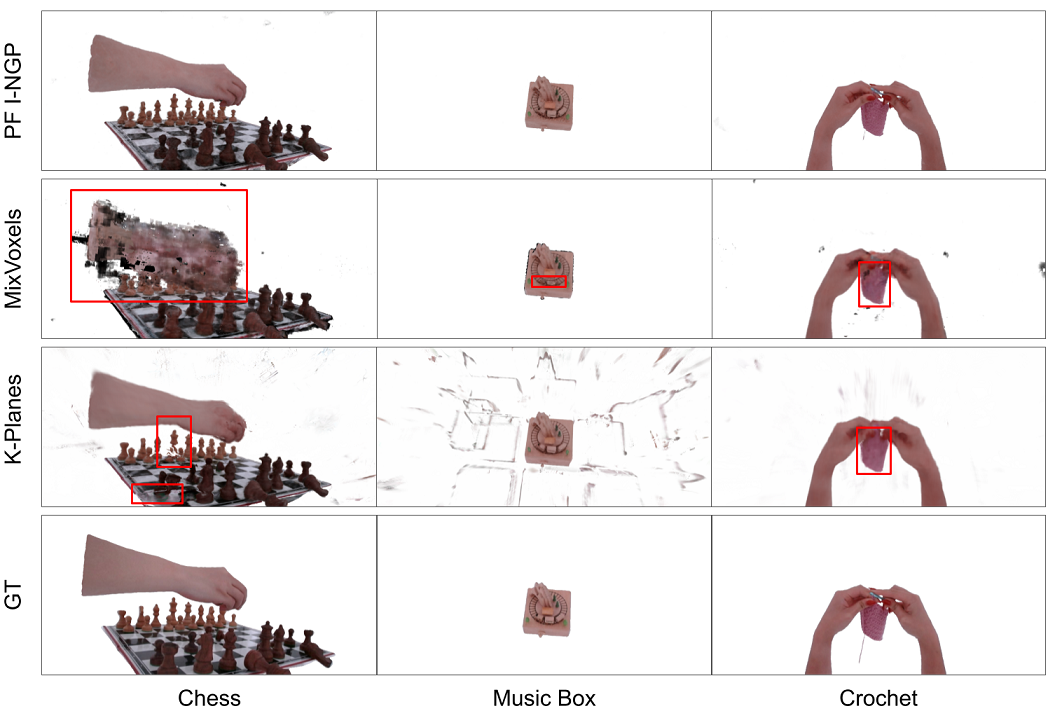}
  \end{subfigure}
  
  \caption{Failure cases of MixVoxels and K-Planes. MixVoxels constructs many floaters around the hand in Chess, fails to reconstruct the small train in Music Box, and the holes of knitted fabric in Crochet. Hence, MixVoxels struggles to capture dynamic parts with complex motion and fine-grained details. K-Planes misses some static parts of Chess, constructs many floaters around objects, especially for slow motion objects such as Music Box, and fills the holes of knitted fabric in Crochet. Therefore, K-Planes struggles to capture static parts and fine-grained details.}
  \label{fig:fail_exp}
\end{figure*}

\parahead{Multi-view 360$^\circ$ Capture System}
In \Cref{fig:exp_g_1,fig:exp_g_2}, PF I-NGP, MixVoxels, and K-Planes can reconstruct World Globe and Wolf better with \textit{All-view} than \textit{Multi-view}.
This indicates that the number of cameras covering the bounding space is significant.
PF I-NGP, MixVoxels, and K-Planes can reconstruct World Globe and Wolf better with \textit{All-view} and \textit{Multi-view} than \textit{Forward} regarding the occluded view.
Hence, 360$^\circ$ setting is necessary to reconstruct the objects with the correct color completely.
Through the visualization results of \textit{Forward}, we observe that MixVoxels tends to render unknown occluded parts with a black background color, while PF I-NGP and K-Planes show better novel view synthesis ability for occluded view.

\section{Foreground-Background Segmentation Method}
As mentioned in Section 4 of the paper, we use I-NGP for foreground-background segmentation and compare the I-NGP method to the Segment Anything (SAM) \cite{kirillov2023segany}. 
In this section, we show the strength of the I-NGP segmentation method and its failure case through visualization.

\begin{figure*}[t!]
  \centering
  \begin{subfigure}[b]{\textwidth}
    \includegraphics[width=\textwidth]{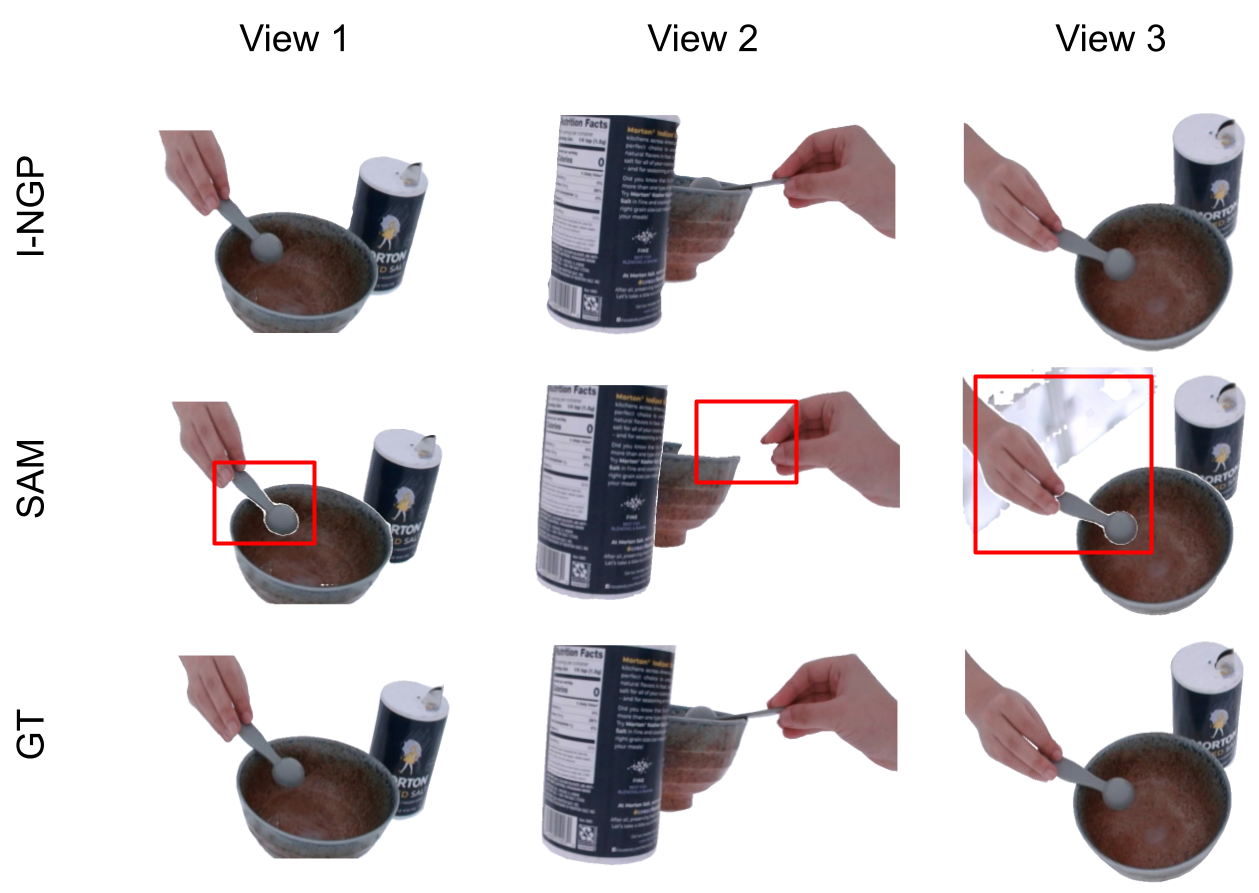}
  \end{subfigure}
  
  \caption{The segmented images of I-NGP and SAM from Pour Salt. SAM cannot maintain multiview consistency, so it contains different artifacts across views. SAM misses the boundary of the spoon in the first view, removes the whole spoon in the second view, and keeps the background in the third view.}
  \label{fig:seg_1}
\end{figure*}

\begin{figure*}[t!]
  \centering
  \begin{subfigure}[b]{\textwidth}
    \includegraphics[width=\textwidth]{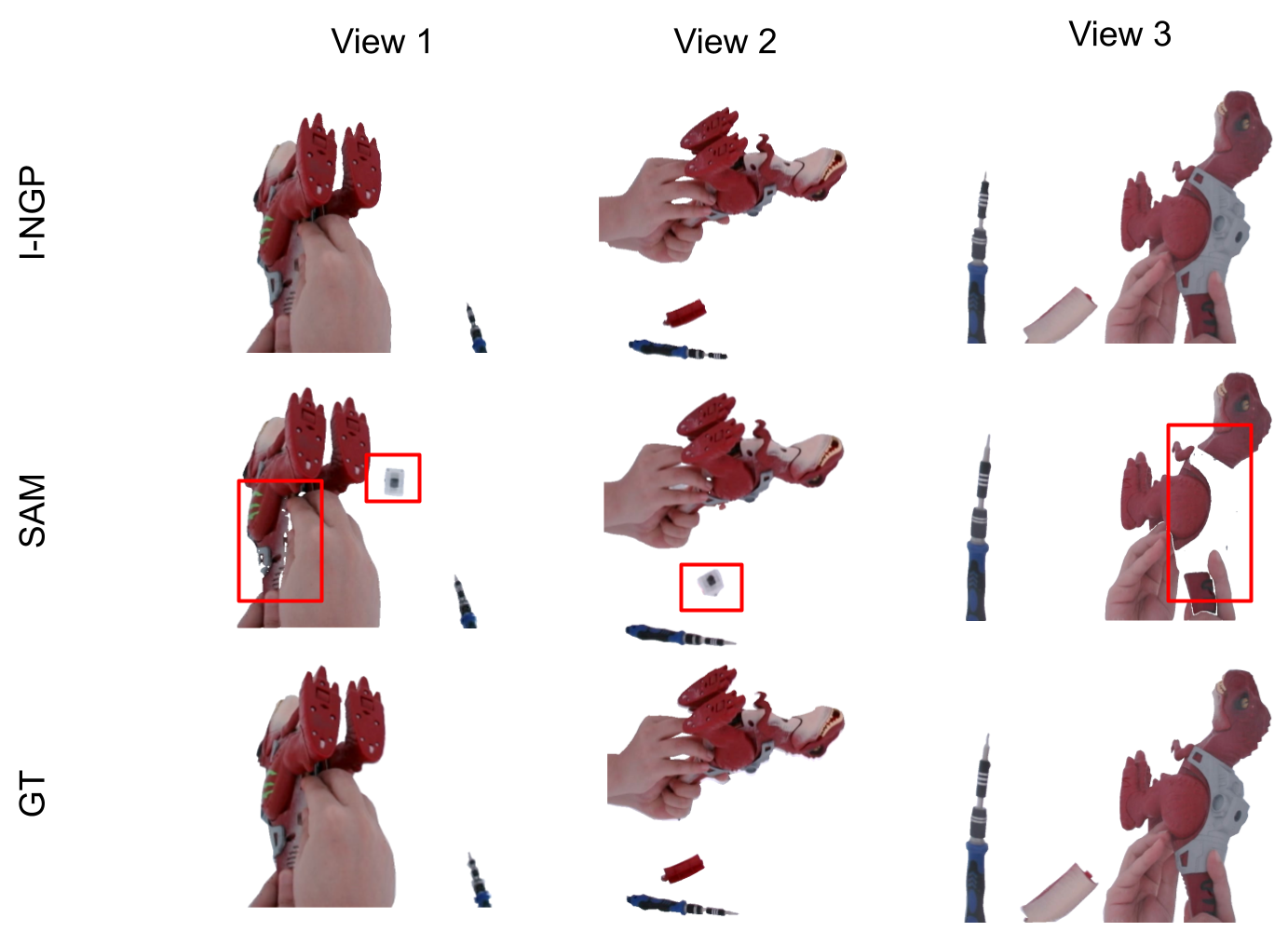}
  \end{subfigure}
  
  \caption{The segmented images of I-NGP and SAM from Replace Battery. SAM cannot maintain multiview consistency, so it contains different artifacts across views. SAM misses the boundary of the hand in the first view, keeps the background in the first and second view, and removes the saddle in the third view.}
  \label{fig:seg_2}
\end{figure*}

\parahead{I-NGP Seg. vs SAM}
We show multi-view inconsistencies of SAM in \Cref{fig:seg_1} and \Cref{fig:seg_2}. 
Specifically, we can notice that different views of the same sequence may contain different artifacts, remove certain objects, or miss the boundary of certain objects. 
For one time step of the Pour Salt sequence, one view contains part of the background, one misses the boundary of the spoon, and one completely removes the spoon. 
Likewise, in the Replace Battery sequence, one view misses the boundary of the hand, one completely removes part of the object, and a few views include the camera in the background. 

\begin{figure*}[t!]
  \centering
  \begin{subfigure}[b]{\textwidth}
    \includegraphics[width=\textwidth]{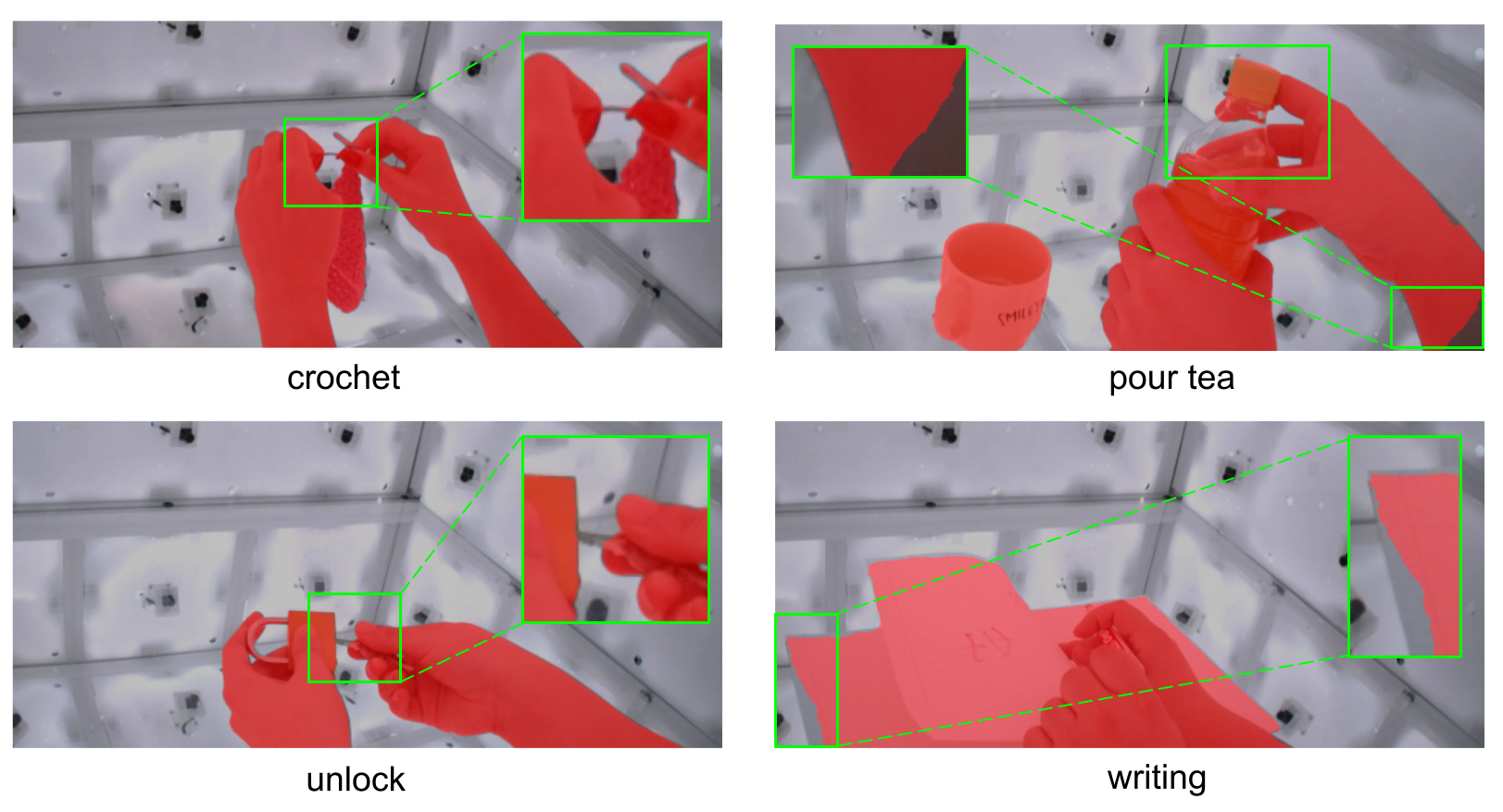}
  \end{subfigure}
  
  \caption{Failure cases of I-NGP segmentation. The performance of I-NGP segmentation is less robust with small thin objects (the yarn in Crochet), transparent objects (the bottle in Pour Tea), high reflection objects (the key in the Unlock), and white objects (the paper in Writing). A part of the lower arm is cut by the bounding box of I-NGP.}
  \label{fig:failed_seg}
\end{figure*}

\begin{figure*}[t!]
  \centering
  \scalebox{0.9}{
  \begin{subfigure}[b]{\textwidth}
    \includegraphics[width=\textwidth]{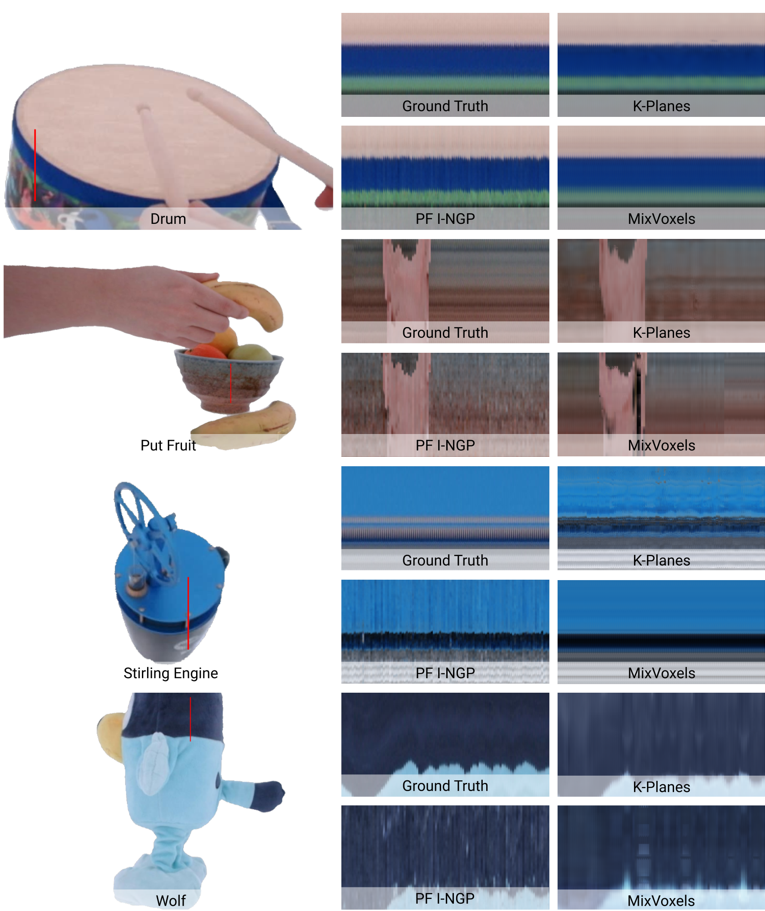}
  \end{subfigure}
  }
  \caption{Visualization of temporal consistency by concatenating a line of pixels across frames from the same view. PF I-NGP contains white noise in all four objects, so PF I-NGP is less temporal consistent. Although K-Planes's rendering result is whiter than the ground truth in Put Fruit, Stirling Engine, and Wolf, the noise is smooth across frames. The rendering result of MixVoxels is pretty smooth in Chess and Stirling Engine, but MixVoxels shows black noise on the hand of the Put Fruit and white blocks in Wolf. }
  \label{fig:tmp_inc}
\end{figure*}

\parahead{Failure Cases} 
We show failure cases of I-NGP segmentation in \Cref{fig:failed_seg}. 
Generally, the method misses sections where the object is very small (e.g. the strand of yarn in the Crochet sequence), transparent (e.g. the clear water bottle in the Pour Tea Sequence), highly reflective (e.g. the key in the Unlock sequence), or white (e.g. the paper in the Writing sequence). 
This is likely because the objects are too similar to the background, making it hard to distinguish between the two. 
It is also important to note that the lower arm in Pour Tea is missed because we shrink the bounding box slightly smaller than the \brics machine. 
This is acceptable because the lower arm is cropped in the 3D space, so it still maintains the multi-view consistency, and we focus on the interaction between objects and hands. 
This also causes the performance of I-NGP segmentation to be underestimated because we label the whole arms for ground truth. 
Of course, the segmentation model in our pipeline can always be replaced by another much better method in the future.

\begin{table*}[tbh!]
  \centering
  \scalebox{0.77}{
  \begin{tabular}{cccccc}
    \toprule
    Scene & PF I-NGP / Mixvoxels / K-Plane & PF I-NGP / Mixvoxels /K-Planes & PF I-NGP / Mixvoxels / K-Planes & PF I-NGP / Mixvoxels / K-Planes \\& PSNR$\uparrow$ & SSIM$\uparrow$ & LPIPS$\downarrow$ & JOD$\uparrow$\\
    \midrule
     chess long & 28.555 / 29.972 / 23.216 & 0.901 / 0.930 / 0.885 & 0.138 / 0.132 / 0.251 & 7.062 / 7.558 / 6.820 \\
     crochet & 29.719 / 28.672 / 26.609 & 0.964 / 0.946 / 0.948 & 0.065 / 0.095 / 0.178 & 8.146 / 7.930 / 6.390 \\
     jenga long & 30.052 / 30.548 / 29.061 & 0.928 / 0.939 / 0.931 & 0.085 / 0.083 / 0.165 &  8.135 / 8.435 / 8.437\\
     legos & 27.927 / 28.436 / 25.342 & 0.887 / 0.901 / 0.890 & 0.132 / 0.122 / 0.212 & 6.514 / 7.492 / 6.611 \\
     origami & 29.694 / 30.191 / 28.028 & 0.933 / 0.936 / 0.932 & 0.095 / 0.094 / 0.175 & 7.796 / 7.955 / 7.740 \\
     painting & 29.937 / 30.151 / 27.881 & 0.921 / 0.933 / 0.922 & 0.118 / 0.112 / 0.292 & 7.557 / 7.851 / 7.316 \\
     puzzle & 28.463 / 29.343 / 25.756 & 0.927 / 0.942 / 0.925 & 0.115 / 0.110 / 0.204 & 7.536 / 8.000 / 7.751 \\
     rubiks cube & 30.436 / 29.649 / 25.668 & 0.969 / 0.951 / 0.937 & 0.064 / 0.093 / 0.209 & 8.024 / 7.776 / 7.112 \\
    \bottomrule\\
  \end{tabular}
  }
  \caption{Rendering quality of all long-duration scenes using PF I-NGP\cite{mueller2022instant}, MixVoxels\cite{wang2023mixed} and K-Planes\cite{kplanes_2023}. It is surprising that MixVoxels outperforms PF I-NGP in most scenes, with the exception of Crochet and Rubik's Cube. This is because PF I-NGP cannot maintain the temporal consistency for static parts, such as the large chessboard in the Chess Long scene and tray in Jenga Long, Legos, Origami, Painting, and Puzzle scenes, while MixVoxels is good at them.}
    \label{tab:dynamic_benchmark_long}
\end{table*}

\section{Dynamic Dataset Distribution}

In this section, we elaborate on the dataset distribution of \shortname.

\begin{figure*}[t!]
  \centering
  \scalebox{0.9}{
  \begin{subfigure}[b]{\textwidth}
    \includegraphics[width=\textwidth]{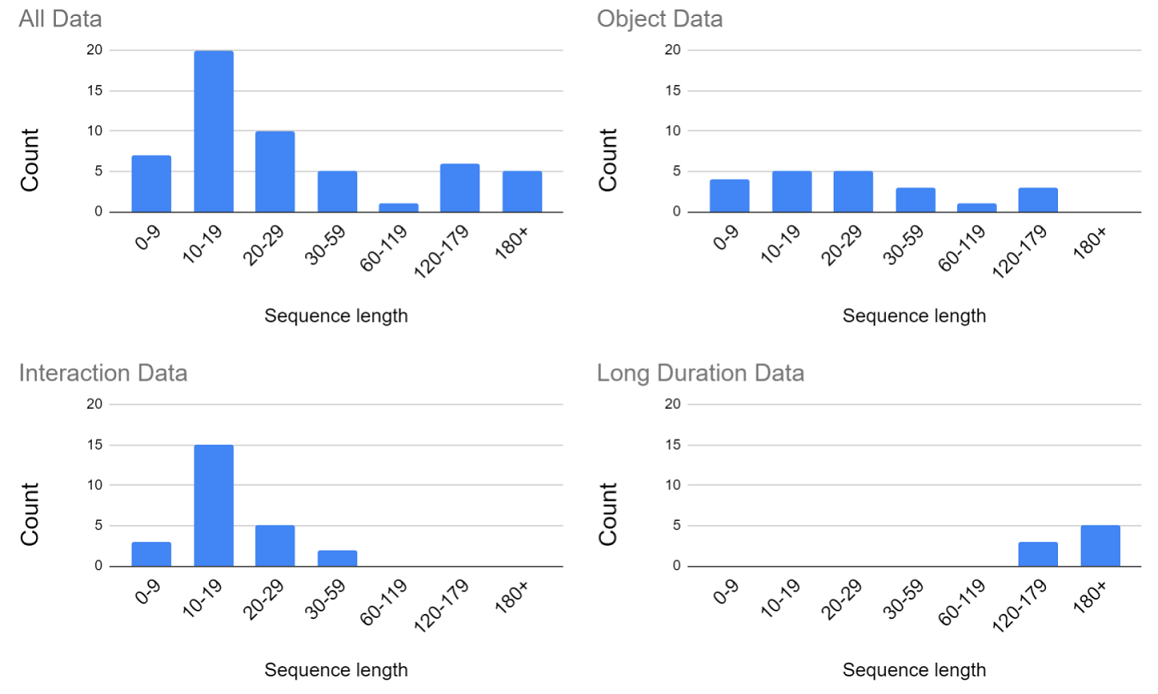}
  \end{subfigure}
  }
  \caption{Data distribution in terms of sequence length. Overall, the sequence length of \shortname ranges from 5 to 200 seconds. Both object and interaction datasets have included several long sequences longer than 10 seconds. Our long-duration dataset provides sequences that are at least 120 seconds long.}
  \label{fig:dist}
\end{figure*}

\parahead{Temporal Length}
\Cref{fig:dist} demonstrates the dataset distribution of all data, object data, interaction data, and long-duration data.
Overall, \shortname contains dynamic sequences ranging from 5 to 200 seconds.
Among them, 37 sequences are 5-29 seconds long, 5 sequences are 30-59 seconds long, 1 sequence is 60-119 seconds long, and 11 sequences are 120-200 seconds long.
Our object and interaction dataset contains several long sequences longer than 30 seconds, and our long-duration dataset contains 8 sequences longer than 119 seconds.

\section{Other metadata}

How can a video play without any audio and captions? 
Hence, our sequences are accompanied by audio captured through microphones and text descriptions labeled by humans for a better viewing experience.
Currently, the audio and text descriptions are only used for a better immersive experience when viewing the sequences. However, we hope this can be extended to other multimodal NeRFs~\cite{liang2023av} in the future.

\parahead{Audio Data} 
\brics can also act as a multimodality capture system that captures visual and sound data simultaneously.
The object sequences and hand-object interaction sequences of the dynamic dataset are accompanied by synchronized spatial audio. 
The 6 microphones located throughout the capture system allow for 360\textdegree audio which provides both loud (e.g. microphones located at the top) and more subtle sounds (e.g. microphones located at the bottom) of the motion. 
We do not currently have synchronized audio for the long-duration dynamic sequences. 

\parahead{Text Description} The object sequences and the hand-object interaction sequences of the dynamic dataset are accompanied by natural language descriptions at 3 levels of detail. These descriptions are generated entirely by a human annotator without the assistance of any automated tools. The coarsest level aims to capture a broad summary of the scene (``putting candy into a mug''), while finer levels increasingly describe appearance (``...the pieces are in pink, green, orange, and black wrappers...''), relative position (``...candy scattered around a black mug...''), number of hands, audio, and temporal progression. Across the dynamic scenes, the average length of the descriptions is 6.1, 18.4, and 38.7 words for the 3 levels of detail, amounting to a total of 2907 words. We do not currently have natural language descriptions for the long-duration dynamic sequences.
To prevent the bias of labeling, we plan to have more people label text descriptions in the future.

\section{Visualization Quality}
We encourage readers to download the uncompressed data from our website for better visualization quality since some blurriness in images is due to the object’s small size relative to the image. For instance, the horse is a hand-size object while the dog is two times larger. This accounts for differences in observable detail between the sequences. We report object sizes in the Table.~\ref{tab:dynamic_object_types}.

\end{document}